\documentclass{article}

\usepackage{microtype}
\usepackage{graphicx}
\usepackage{amsmath}
\usepackage{amssymb}
\usepackage{booktabs} 
\usepackage{listings}
\usepackage[font=small]{./subcaption}
\usepackage{./subfiles}

\usepackage{hyperref}

\usepackage[accepted]{./icml2019}
\usepackage{./algpseudocode}

\begin{document}

\twocolumn[
\icmltitle{Mitigating large adversarial perturbations on X-MAS (X minus Moving Averaged Samples)}



\icmlsetsymbol{equal}{*}

\begin{icmlauthorlist}
\icmlauthor{Woohyung Chun}{samsung}
\icmlauthor{Sung-Min Hong}{samsung}
\icmlauthor{Junho Huh}{samsung}
\icmlauthor{Inyup Kang}{samsung}
\end{icmlauthorlist}

\icmlaffiliation{samsung}{Samsung Electronics System LSI Division, Hwaseong-si, Gyeonggi-do, South Korea}
\icmlcorrespondingauthor{Woohyung Chun}{wh.chun@samsung.com}

\icmlkeywords{Trustworthy Machine Learning, robust inference, adversarial perturbation, ICML}

\vskip 0.3in
]



\printAffiliationsAndNotice{}  

\begin{abstract}
We propose the scheme that mitigates the adversarial perturbation $\epsilon$ on the adversarial example $X_{adv}$ ($=$ $X$ $\pm$ $\epsilon$, $X$ is a benign sample) by subtracting the estimated perturbation $\hat{\epsilon}$ from $X$ $+$ $\epsilon$ and adding $\hat{\epsilon}$ to $X$ $-$ $\epsilon$.
The estimated perturbation $\hat{\epsilon}$ comes from the difference between $X_{adv}$ and its moving-averaged outcome $W_{avg}*X_{adv}$ where $W_{avg}$ is $N \times N$ moving average kernel that all the coefficients are one.
Usually, the adjacent samples of an image are close to each other such that we can let $X$ $\approx$ $W_{avg}*X$ (naming this relation after X-MAS[X minus Moving Averaged Samples]).
By doing that, we can make the estimated perturbation $\hat{\epsilon}$ falls within the range of $\epsilon$.
The scheme is also extended to do the multi-level mitigation by configuring the mitigated adversarial example $X_{adv}$ $\pm$ $\hat{\epsilon}$ as a new adversarial example to be mitigated.
The multi-level mitigation gets $X_{adv}$ closer to $X$ with a smaller (i.e. mitigated) perturbation than original unmitigated perturbation by setting the moving averaged adversarial sample $W_{avg} * X_{adv}$ (which has the smaller perturbation than $X_{adv}$ if $X$ $\approx$ $W_{avg}*X$) as the boundary condition that the multi-level mitigation cannot cross over (i.e. decreasing $\epsilon$ cannot go below and increasing $\epsilon$ cannot go beyond).
With the multi-level mitigation, we can get high prediction accuracies even in the adversarial example having a large perturbation (i.e. $\epsilon$ $>$ $16$).
The proposed scheme is evaluated with adversarial examples crafted by the FGSM (Fast Gradient Sign Method) based attacks on ResNet-50 trained with ImageNet dataset.
\end{abstract}

\section{Introduction} \label{sec:intro}
Adversarial perturbations~\cite{alexey, cnw, chuan} for a CNN (Convolutional Neural Network) are crafted to make a CNN classifier mispredict its input image. 
Usually, perturbations are imperceptible to human eyes especially when their size(as known as $\epsilon$)s are small (i.e. $\epsilon$ $\leq$ $16$ in the FGSM [Fast Gradient Sign Method] based attack~\cite{alexey}).
However, even in the case that an adversarial perturbation on an input image is large ($>$ 16), human beings can classify the image with a correct label.

For the defense of a CNN against the adversarially perturbed images, there have been the approaches~\cite{thermometer, countering, distillation, scale} to make neural networks robust against the perturbations. 
Fundamentally, the approaches need to have a full data set of neural networks for a re-training. 
Also, there are some other approaches which do not require any modification of neural networks~\cite{alexey,gintare,deflecting}.
The approaches utilize the property that adversarial perturbations tend to exist on a high-frequency region~\cite{chuan,yash}. 
Thus, the low-pass image filter (i.e. DCT [Discrete Cosine Transform] or wavelet denoising) deletes the perturbations on a high-frequency region by cutting off high-frequency components.
However, they are not so effective for the adversarial example having a large perturbation~\cite{gintare}.

The image transformations such as DCT and wavelet were developed to make the image with less information (i.e. compressed one with less bits) seemingly have a good quality comparable to original one for human beings. 
They utilize that human beings are not sensitive to small changes of an image.
If the techniques originally developed for the less-sensitive human beings (not for the defense against the adversarial perturbations) can be used to invalidate a relatively small perturbation (i.e. $\epsilon$ $\leq$ $16$), it would be able to nullify a large perturbation if we can make the large perturbation become a smaller one that the image transformation can delete.

For the mitigation of an adversarial perturbation $\epsilon$ of the adversarial example $X_{adv}$ ($=$ $X$ $\pm$ $\epsilon$),
we find the estimated perturbation $\hat{\epsilon}$ and add it to $X_{adv}$ in the direction of making $|\epsilon|$ small (i.e. $X_{adv}$ $-$ $\hat{\epsilon}$ for $+$ $\epsilon$, $X_{adv}$ $+$ $\hat{\epsilon}$ for $-$ $\epsilon$).
The estimated perturbation comes from the difference between $X_{adv}$ and its moving-averaged outcome $W_{avg}*X_{adv}$ where $W_{avg}$ is $N \times N$ moving average kernel that all the coefficients are one.
Since $X$ $\approx$ $W_{avg}*X$ among the adjacent samples of an image, we can make the difference between $X_{adv}$ and $W_{avg}*X_{adv}$ less than $\epsilon$.
For a large adversarial perturbation (e.g. $\epsilon$ $>$ $16$), the mitigation scheme is extended to do the multi-level mitigation by configuring the mitigated adversarial example as a new adversarial example to be mitigated further.
Also, in order to guarantee that more mitigation steps get the adversarial perturbation smaller, the less perturbed moving averaged sample (i.e. $W_{avg}*X_{adv}$ is closer to $X$ than $X_{adv}$ if $X$ $\approx$ $W_{avg}*X$)
is set as the boundary condition that the mitigation by subtraction ($X_{adv}$ $-$ $\hat{\epsilon}$) does not go below and the mitigation by addition ($X_{adv}$ $+$ $\hat{\epsilon}$) does not go beyond.

This paper is organized as follows:
Section~\ref{sec:problem_setup} sets up the function $M(\cdot)$ which mitigates a large perturbation that JPEG encoding cannot nullify. 
$M(\cdot)$ is dedicated to making the large perturbation small such that JPEG encoding is separated as the soothing function from $M(\cdot)$.
In Section~\ref{sec:proposed_method}, the way of estimating perturbation is introduced on X-MAS (i.e. when $X$ $\approx$ $W_{avg}*X$) and it is extended to the multilevel-mitigation for a large perturbation. In the end of the section, the proposed mitigation schemes are built in an algorithm with some relevant parameters.
Section~\ref{sec:evaluation} evaluates the algorithm with some representing adversarial examples having large perturbations.
Finally, Section~\ref{sec:conclusion} concludes with the summary of our contribution.

\section{Problem Setup} \label{sec:problem_setup}
\begin{figure}[htb]
\begin{subfigure}[htb]{0.4\textwidth}
	\centering
  \includegraphics[width=\textwidth]{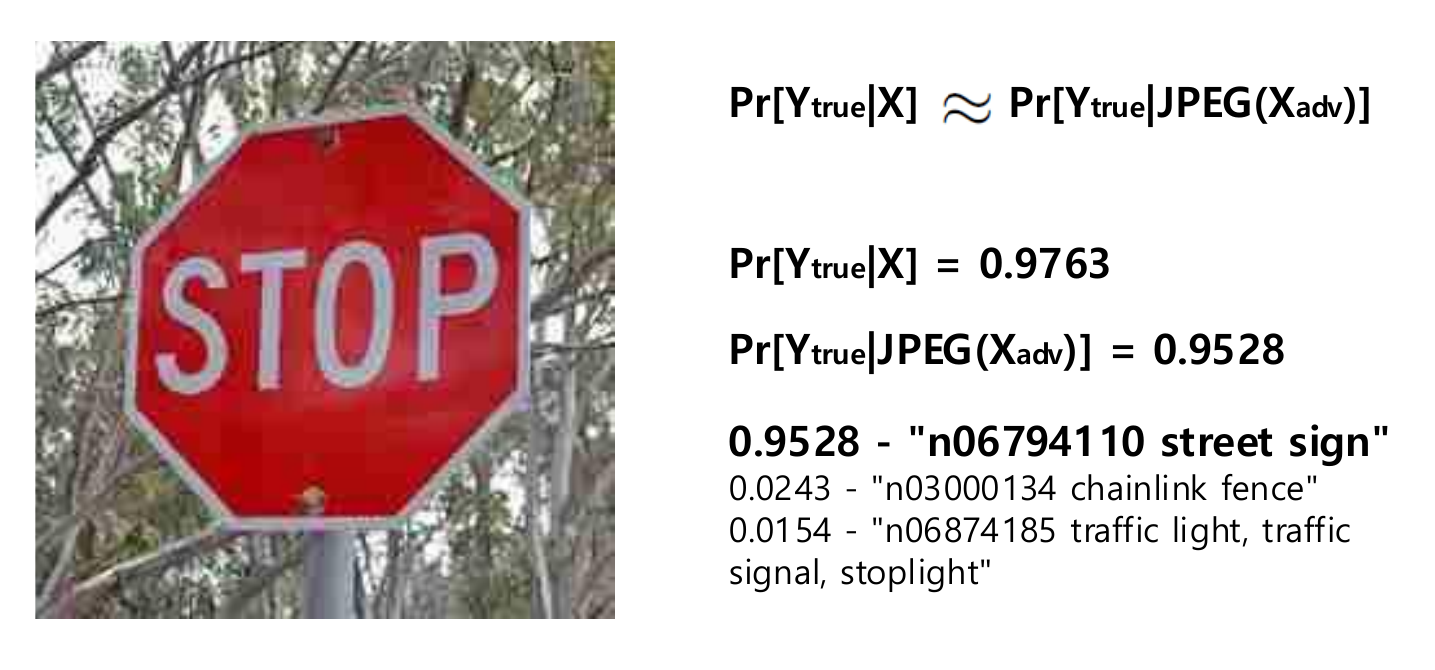}
  \caption{JPEG encoding can well soothe the small perturbations generated by the basic iterative FGSM attack with $\epsilon$ $=$ $2$.}
  \label{fig:jpeg_good}
\end{subfigure}
\begin{subfigure}[htb]{0.4\textwidth}
	\centering
  \includegraphics[width=\textwidth]{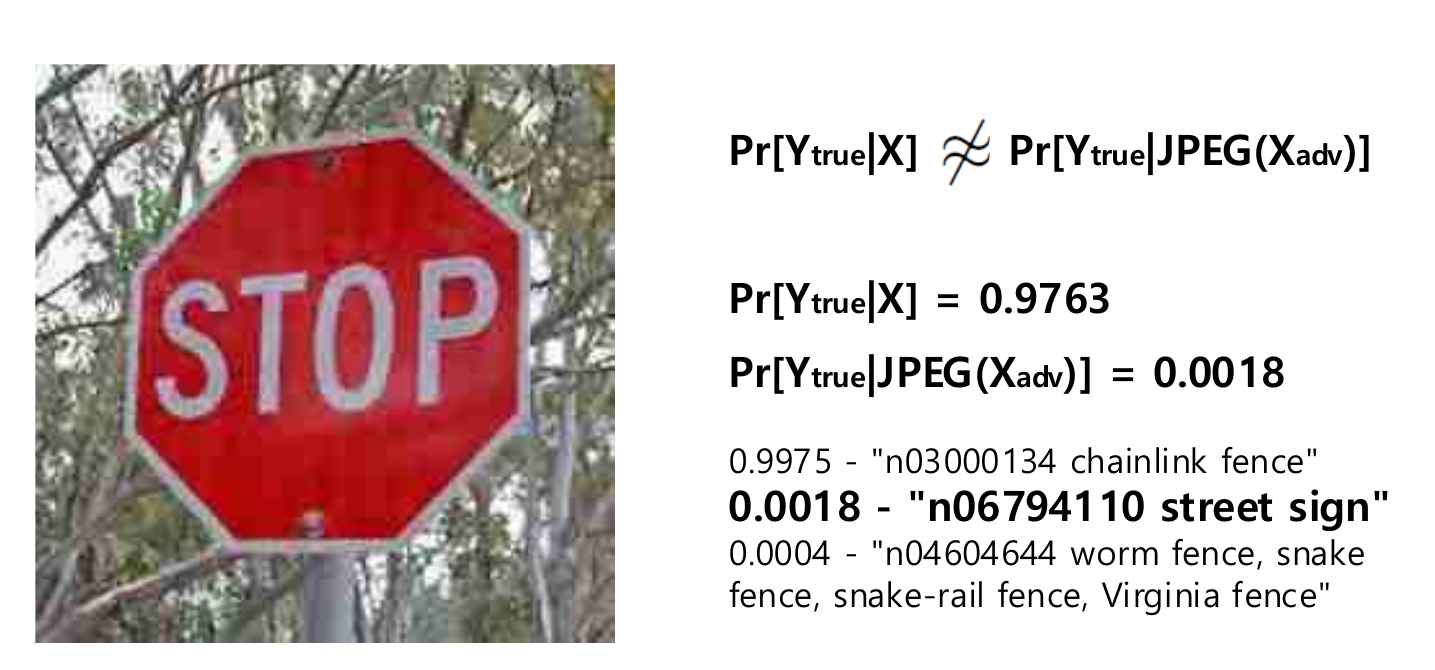}
  \caption{JPEG encoding cannot well soothe the large perturbations crafted by the basic iterative FGSM attack with $\epsilon$ $=$ $32$. }
  \label{fig:jpeg_bad}
\end{subfigure}
\caption{JPEG encoding does not work well for a large perturbation.}
\label{fig:jpeg_soothe}
\end{figure}

Previous researches~\cite{alexey, gintare, chuan, yash} identified that JPEG encoding is able to nullify the impact of the adversarial perturbations on the prediction accuracy of a CNN.
When the adversarial example $X_{adv}$ is JPEG-encoded and then fed into a CNN which recognizes the benign version of $X_{adv}$, $X$ as $Y_{true}$,
\begin{align}
  Pr[Y_{true}|X] &\approx Pr[Y_{true}|JPEG(X_{adv})], \nonumber \\
  X_{adv} &= X \pm \epsilon, \;\; \epsilon \geq 0 \nonumber \\
\label{eqn:jpeg_soothe}
\end{align}
where $Pr[Y_{true}|X]$ is the accuracy that a CNN predicts $X$ as the true label $Y_{true}$ and $\epsilon$ is the adversarial perturbation applied to all the samples of an input image for a CNN.
Usually, the relation "$\approx$" of Equation (\ref{eqn:jpeg_soothe}) works well for the small perturbations but it is not valid for the large perturbations as shown in Figure \ref{fig:jpeg_soothe}.

In Figure \ref{fig:jpeg_soothe}, the perturbations $\epsilon$s are crafted by the basic iterative FGSM attack~\cite{alexey}.
Figure \ref{fig:jpeg_good} satisfies the relation "$\approx$" of Equation (\ref{eqn:jpeg_soothe}) but Figure \ref{fig:jpeg_bad} does not.
$JPEG(\cdot)$ compresses the image with the quality of 20 (out of 100).
In order to make Figure \ref{fig:jpeg_bad} have the prediction accuracy comparable to  Figure \ref{fig:jpeg_good}, we need to mitigate $X_{adv}$ of Figure \ref{fig:jpeg_bad} to the level of $X_{adv}$ in Figure \ref{fig:jpeg_good}.
Let the mitigation function $M(\cdot)$ that mitigates the adversarial perturbation $\epsilon$ of Equation (\ref{eqn:jpeg_soothe}) be
\begin{align}
  M(X_{adv}) &= X_{adv} \pm \hat{\epsilon} \nonumber \\
             &= X \pm \epsilon \mp \hat{\epsilon}, \;\; 0 \leq \hat{\epsilon} \leq \epsilon \nonumber \\
  \label{eqn:def_miti}
\end{align}
where $\hat{\epsilon}$ is the estimated perturbation which mitigates the perturbation by subtracting $\hat{\epsilon}$ for $\epsilon$ $>$ 0 and adding $\hat{\epsilon}$ for $\epsilon$ $<$ 0.
To make the difference between $M(X_{adv})$ and $X_{adv}$ be imperceptible to human eyes, $\hat{\epsilon}$ must be within the range of $\epsilon$. 
That is, $\hat{\epsilon}$ should work on the direction of decreasing the range of $\epsilon$ from [$-\epsilon$,  $+\epsilon$] to [$-\epsilon$ $+$ $\hat{\epsilon}$,  $+\epsilon$ $-$ $\hat{\epsilon}$] where $0$ $\leq$ $\hat{\epsilon}$ $\leq$ $\epsilon$.
In Equation (\ref{eqn:def_miti}), $\hat{\epsilon}$ is subtracted for $X_{adv}$ $=$ $X$ $+$ $\epsilon$
and it is added to $X_{adv}$ $=$ $X$ $-$ $\epsilon$.
By applying the mitigation function $M(\cdot)$, Equation (\ref{eqn:jpeg_soothe}) can be expressed with the terms as below
\begin{align}
  Pr[Y_{true}|X] &\approx Pr[Y_{true}|SF(M(X_{adv}))], \nonumber \\
  X_{adv} &= X \pm \epsilon, \nonumber \\
  M(X_{adv}) &= X_{adv} \pm \hat{\epsilon}, \;\; 0 \leq \hat{\epsilon} \leq \epsilon \nonumber \\
  \label{eqn:def_soothe}
\end{align}
where $SF(\cdot)$ is the soothing function corresponding to JPEG encoding in Equation (\ref{eqn:jpeg_soothe}).
Soothing function reduces the impact of the perturbation on the prediction accuracy.
As a soothing function, JPEG encoding removes the high-frequency perturbations through DCT (Discrete Cosine Transform) and quantization~\cite{chuan,yash}. 
Also, the simple moving average filter (which runs the convolutional computations with the weight kernel where all the coefficients are one) can be used as the soothing function by smoothing the perturbations through the spatial average.
In Section \ref{sec:proposed_method}, we propose the method of estimating $\hat{\epsilon}$ in Equation (\ref{eqn:def_miti}) and the way of mitigating the adversarial perturbation using $\hat{\epsilon}$.

\section{Proposed method} \label{sec:proposed_method}
\subsection{Estimated perturbation $\hat{\epsilon}$} \label{sec:estimated_perturbation}
In order to find $\hat{\epsilon}$, we use the moving average filters that make $X_{adv}$ converge some value ranging from [X $-$ $\epsilon$, $X$ $+$ $\epsilon$]. That is,
\begin{align}
&W_{avg}*X_{adv} = W_{avg}*X \pm W_{avg}*\epsilon 
\label{eqn:moving_avg}
\end{align}
where $\epsilon$ $\geq$ $0$ and $W_{avg}$ is $N \times N$ (N $\geq$ 2) moving average window whose samples are convolved with the kernel that all the coefficients are one. The moving average operation "$W_{avg}*$" decreases the difference between adjacent samples. Thus, the moving-average operation mitigates the perturbation $\epsilon$ as below.
\begin{align}
&W_{avg}*X_{adv} - W_{avg}*X = \pm W_{avg}*\epsilon, \nonumber \\
&W_{avg}*|\epsilon| < |\epsilon| \;\; \text{    if } ( \frac{1}{|W_{avg}|}\sum_{n=0}^{|W_{avg}|-1}{\epsilon_{n}} ) \neq \epsilon, \nonumber \\
&W_{avg}*\epsilon  = \epsilon \;\; \text{    if } ( \frac{1}{|W_{avg}|}\sum_{n=0}^{|W_{avg}|-1}{\epsilon_{n}} ) = \epsilon
\nonumber \\
\label{eqn:moving_eps}
\end{align}
where $|W_{avg}|$ is the number of coefficients for $N \times N$ $W_{avg}$, $N^{2}$ and $\epsilon_{n}$ is the perturbation assigned to each sample that $W_{avg}$ covers.
In Equation(\ref{eqn:moving_eps}), the cases satisfying $W_{avg}*|\epsilon|$ $<$ $|\epsilon|$ are more probable than the cases that meet $W_{avg}*\epsilon$ $=$ $\epsilon$.
For example, in order to make $W_{avg} * \epsilon$ $=$ $\epsilon$ in FGSM (Fast Gradient Sign Method) attack~\cite{fgsm}, all the samples in the coverage of $W_{avg}$ have $\epsilon$.
The probability that each sample has the same $\epsilon$ in FGSM is $\frac{1}{3}$ (i.e $+\epsilon$ among $-\epsilon$, 0 and $+\epsilon$).
So, the probability that the result of $N \times N$ moving-average computation becomes $\epsilon$ is $\frac{1}{3^{N^2}}$.
Its value is about $5 \times {10}^{-5}$ when $3 \times 3$ kernel is used for $W_{avg}$.
In the same manner, we can find the probability that the result of $N \times N$ moving-average computation becomes $-\epsilon$ and it is also $\frac{1}{3^{N^2}}$.
Thus, when $\epsilon$ is generated by the rule of FGSM and the moving average kernel $W_{avg}$ is $3 \times 3$, the probability that the inequality $W_{avg}*|\epsilon|$ $<$ $|\epsilon|$ happens is $1$ $-$ $\frac{2}{3^{3^2}}$ ($\approx$ 0.9999).
Since it is highly probable that $W_{avg}*|\epsilon|$ $<$ $|\epsilon|$, we can use $|\epsilon|$ $-$ $W_{avg}*|\epsilon|$ as the estimated perturbation $\hat{\epsilon}$ in order to make $\epsilon$ smaller.
When $X$ $\approx$ $W_{avg}*X$, $\hat{\epsilon}$ can be found by
\begin{align}
\forall X_{adv} &> W_{avg}*X_{adv}, \nonumber \\
\hat{\epsilon} &= X_{adv} - (W_{avg}*X_{adv}) \nonumber \\
               &= X + \epsilon - W_{avg}*(X + \epsilon) \nonumber \\
               &\approx \epsilon - W_{avg}*\epsilon \;\; (\because X \approx W_{avg}*X), \nonumber \\
	       \nonumber \\
\forall X_{adv} &< W_{avg}*X_{adv} \nonumber, \\
\hat{\epsilon} &= (W_{avg}*X_{adv}) - X_{adv} \nonumber \\
               &= W_{avg}*(X - \epsilon) - (X - \epsilon) \nonumber \\
               &\approx \epsilon - W_{avg}*\epsilon \;\; (\because X \approx W_{avg}*X) \nonumber \\
\label{eqn:hat_eps}
\end{align}
where $\epsilon$ $>$ 0, $X_{adv}$ $=$ $X$ $+$ $\epsilon$ if the moving average output of $X_{adv}$ is smaller than $X_{adv}$ (i.e. $X_{adv}$ $>$ $W_{avg}*X_{adv}$) and $X_{adv}$ $=$ $X$ $-$ $\epsilon$ if the moving average output of $X_{adv}$ is larger than $X_{adv}$ ((i.e. $X_{adv}$ $<$ $W_{avg}*X_{adv}$).
$X$ $\approx$ $W_{avg}*X$ can be satisfied by controlling $|W_{avg}|$  
(usually, adjacent samples are close to each other).

\begin{figure}[htb]
  \centering
  \includegraphics[width=1.0\linewidth]{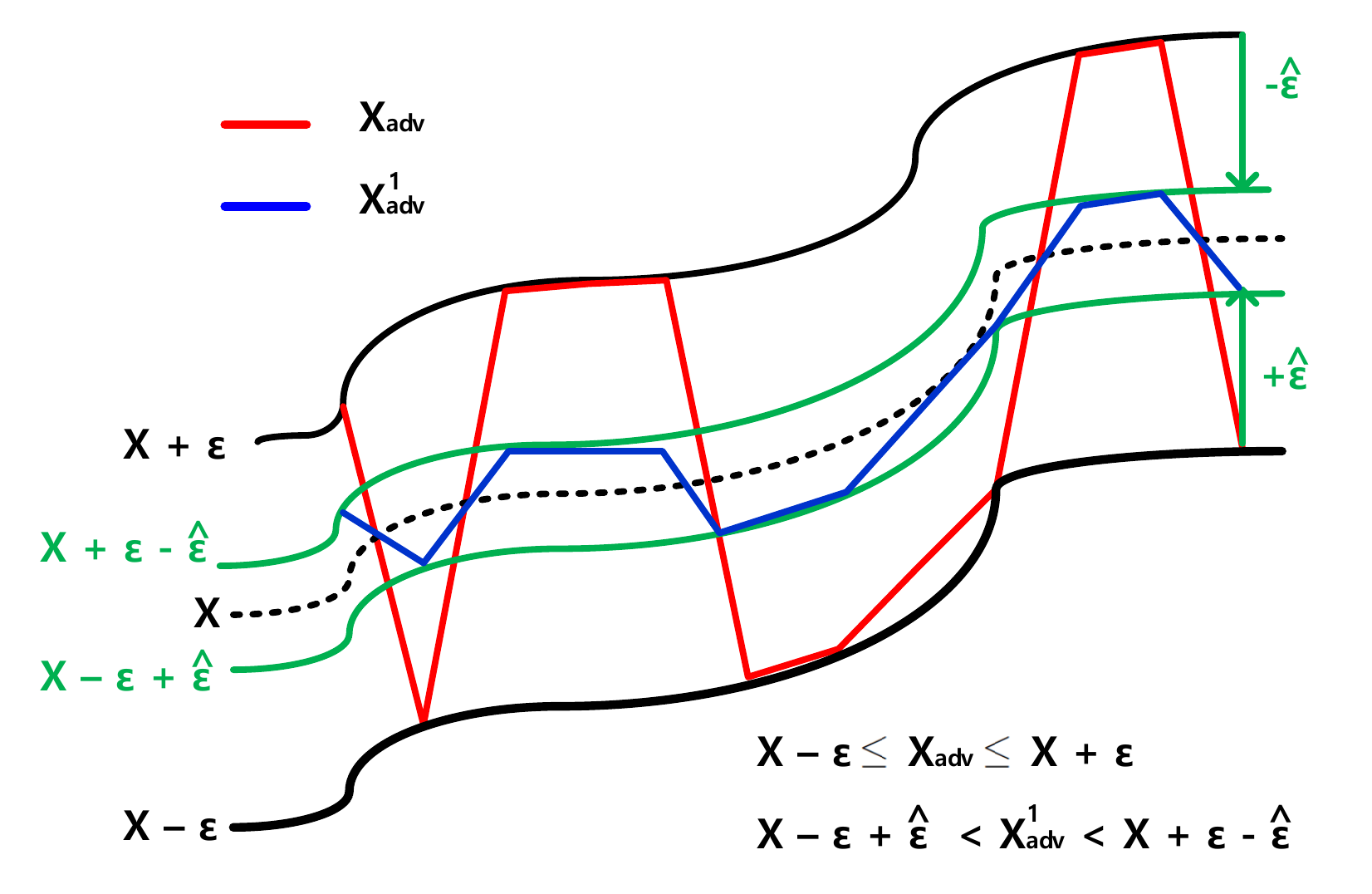}
  \caption{Illustration of mitigating $\epsilon$ with $\hat{\epsilon}$}
  \label{fig:mitigation_with_hat_of_eps}
\end{figure}

In Equation (\ref{eqn:hat_eps}), $0$ $<$ $\hat{\epsilon}$ $<$ $\epsilon$ such that $\hat{\epsilon}$  can mitigate $\epsilon$ by either being subtracted from $\epsilon$ or being added to $\epsilon$ 
That is, when $X_{adv}$ $>$ $W_{avg}*X_{adv}$, $\hat{\epsilon}$ is subtracted from $X_{adv}$ and $\hat{\epsilon}$ is added to $X_{adv}$ if $X_{adv}$ $<$ $W_{avg}*X_{adv}$.
Figure \ref{fig:mitigation_with_hat_of_eps} illustrates how $\hat{\epsilon}$ mitigates $\epsilon$.

In Figure \ref{fig:mitigation_with_hat_of_eps}, the solid black line shows the upper and lower limits where the adversarial perturbation $\epsilon$ works on $X$.
Thus, the red line which denotes the adversarial example $X_{adv}$ having $\pm\epsilon$ as the adversarial perturbation, does not get out of the black line.
Also, the solid green line indicates the upper and lower boundaries that $\hat{\epsilon}$ works on $X_{adv}$.
If ($X$ $+$ $\epsilon$ $-$ $\hat{\epsilon}$) $<$ $X_{adv}$ $<$ ($X$ $+$ $\epsilon$), $X_{adv}$ has the positive $\epsilon$ such that $\hat{\epsilon}$ ($=$ $\epsilon$ $-$ $W_{avg}*\epsilon$)  should be subtracted from $X_{adv}$ to reduce $+\epsilon$.
On the other hand, if ($X$ $-$ $\epsilon$) $<$ $X_{adv}$ $<$ ($X$ $-$ $\epsilon$ $+$ $\hat{\epsilon}$), 
$X_{adv}$ has the negative $\epsilon$. Thus, $\hat{\epsilon}$ should be added into $X_{adv}$ to increase $-\epsilon$.
Both $X_{adv}$ $+$ $\hat{\epsilon}$ and $X_{adv}$ $-$ $\hat{\epsilon}$ are closer to $X$ than $X$ $+$ $\epsilon$ and $X$ $-$ $\epsilon$ are.
In the same manner with $X_{adv}$, the solid blue line that represents the adversarial example $X_{adv}^{1}$ having $\pm(\epsilon - \hat{\epsilon})$ as the adversarial perturbation, does not get out of the green line.
If $\hat{\epsilon}$ $=$ 0 (i.e. $X_{adv}$ $=$ $W_{avg}*X_{adv}$), all the samples within the coverage of $W_{avg}$ have the same $\epsilon$.
As discussed earlier, the probability that all the samples have the same $\epsilon$ is very low even in the small $3 \times 3$ $W_{avg}$.
It is much smaller than the probability that $\epsilon$ of $X_{adv}$ is 0 (e.g. $\frac{1}{3}$ in the FGSM based attack).
Thus, when $\hat{\epsilon}$ $=$ 0, it is more probable that the perturbation of $X_{adv}$ is zero 
rather than that all the samples have the same $\epsilon$.
Since we do not have to find $\hat{\epsilon}$ for the sample having no perturbation, we do not change $X_{adv}$ in case that $\hat{\epsilon}$ $=$ 0.

\begin{figure}[htb]
\begin{subfigure}[htb]{0.4\textwidth}
	\centering
  \includegraphics[width=\textwidth]{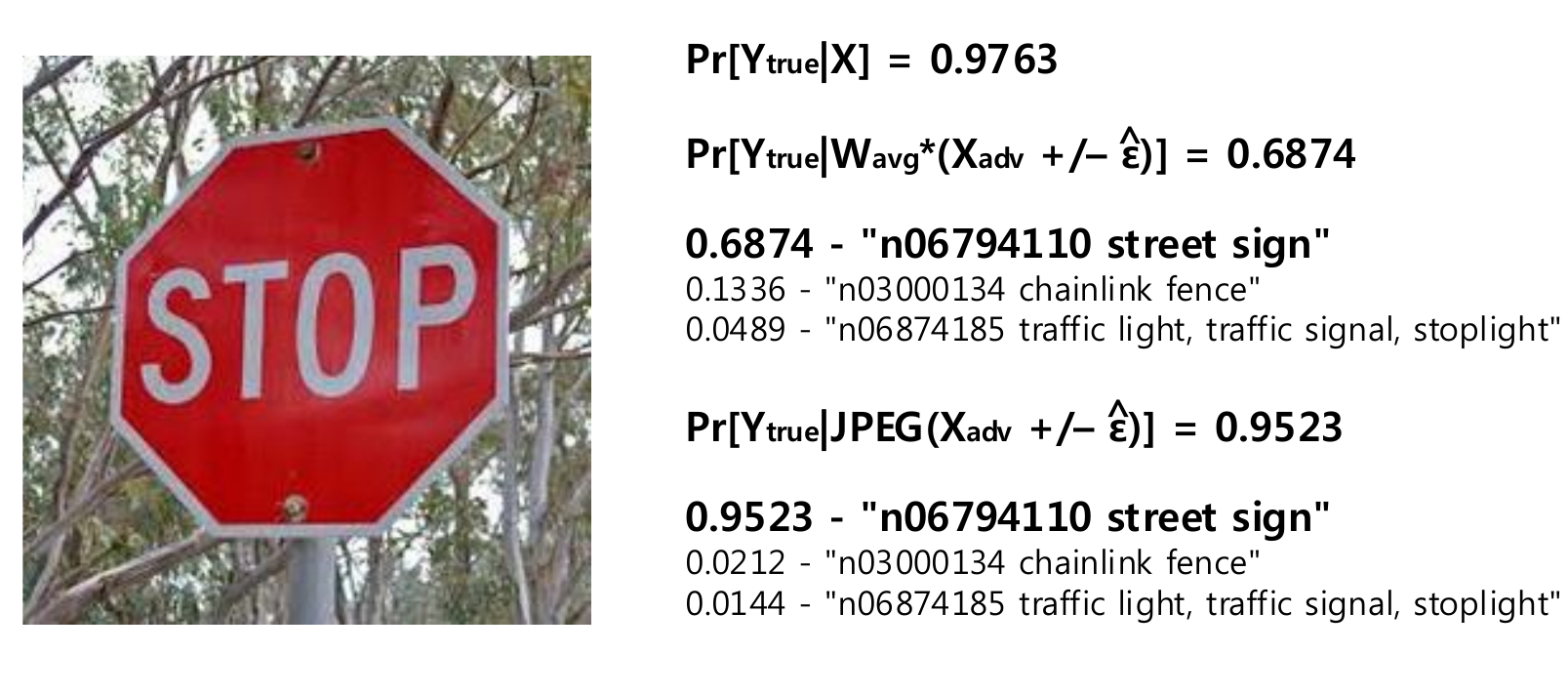}
  \caption{Single-level mitigation can well mitigate the small perturbations crafted by the basic iterative FGSM attack with $\epsilon$ $=$ $2$. }
  \label{fig:eps2_single}
\end{subfigure}
\begin{subfigure}[htb]{0.4\textwidth}
	\centering
  \includegraphics[width=\textwidth]{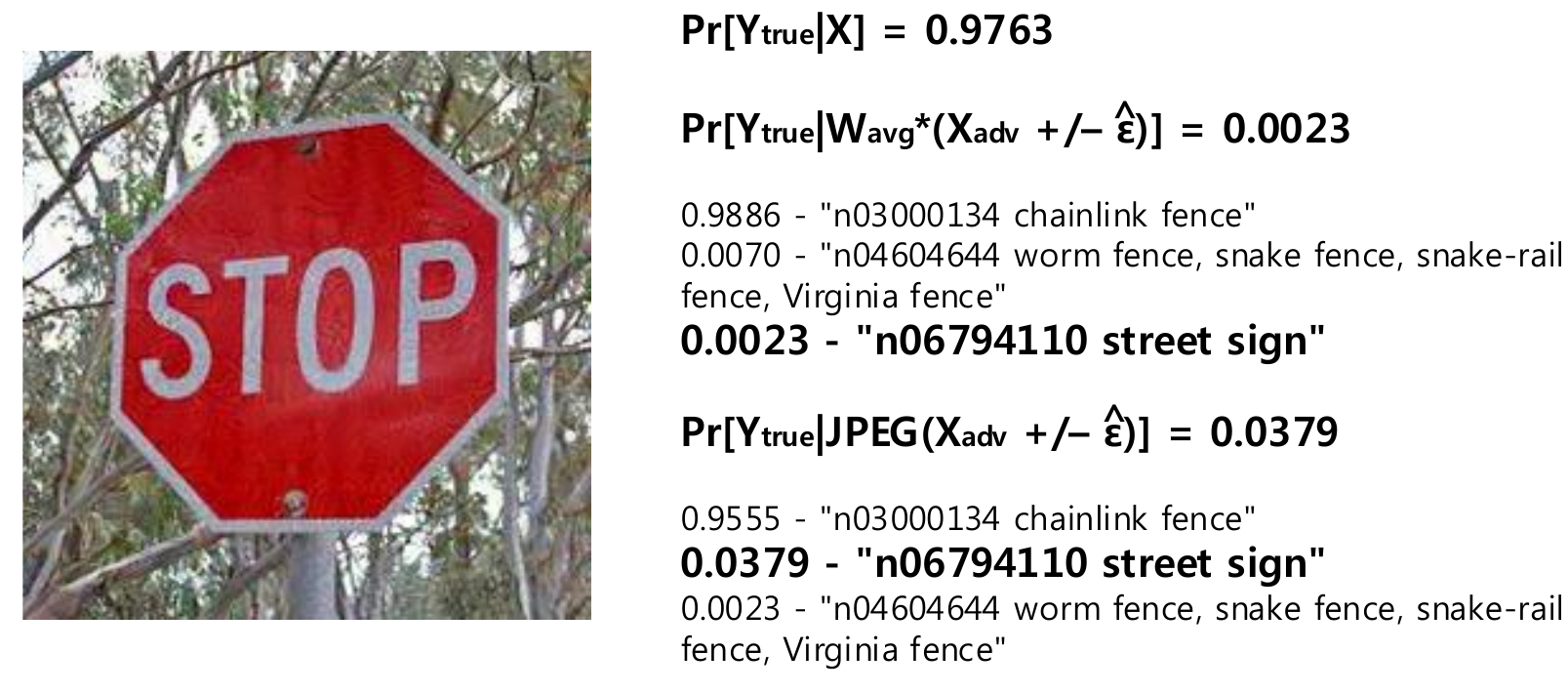}
  \caption{Single-level mitigation may not work well on large perturbations crafted by the basic iterative FGSM attack with $\epsilon$ $=$ $32$.}
  \label{fig:eps32_single}
\end{subfigure}
\caption{Impact of the single-level mitigation on the prediction accuracy according to the sizes of $\epsilon$ in $X_{adv}$}
\label{fig:single_mitigation}
\end{figure}

In Equation (\ref{eqn:hat_eps}), $\hat{\epsilon}$ can be larger than $\epsilon$ if $X$ $\not\approx$ $W_{avg}*X$.
It is the case that samples within the moving average kernel are very different to each other.
Then, $\hat{\epsilon}$ may work as another perturbation.
Also, $\hat{\epsilon}$ can be too small to improve the prediction accuracy $Pr[Y_{true}|(X_{adv}\pm\hat{\epsilon})]$.
For both large and small $\hat{\epsilon}$, more mitigation steps would be required.
Figure \ref{fig:single_mitigation} shows the impact of the single-level mitigation on the prediction accuracy according to the sizes of $\epsilon$ in $X_{adv}$.

In Figure \ref{fig:single_mitigation}, $|W_{avg}|$ $=$ $3$ $\times$ $3$ 
for both $\hat{\epsilon}$ and the soothing filter "$W_{avg}*$" of $Pr[Y_{true}|W_{avg}*(X_{adv} +/- \hat{\epsilon})]$. 
$JPEG(\cdot)$ compresses the image in the quality with 20 (out of 100).
Figure \ref{fig:eps2_single} shows the case where single-level mitigation works very well on the small perturbations to get the high prediction accuracy. 
However, the single-level mitigation is not so effective on the large perturbation in Figure \ref{fig:eps32_single}.
In order to achieve a high prediction accuracy on the adversarial example having large perturbations, we need to run a multi-level mitigation in a controlled manner.

\subsection{Multi-level mitigation} \label{sec:multilevel_mitigation}
\begin{figure}[htb]
  \centering
  \includegraphics[width=1.0\linewidth]{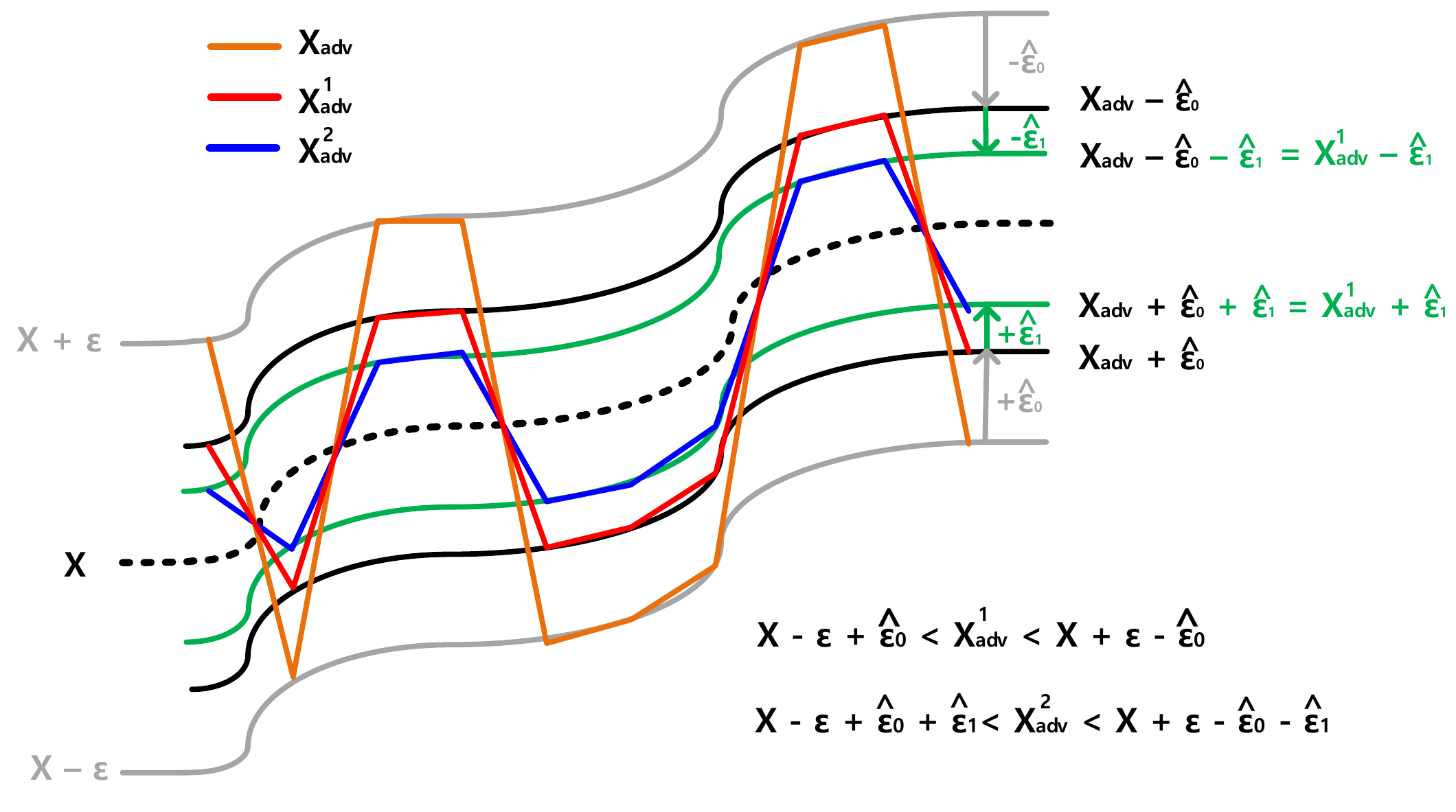}
  \caption{Illustration of the multi-level mitigation}
  \label{fig:multilevel_mitigation_illustration}
\end{figure}

The single-level mitigation with the estimated perturbation $\hat{\epsilon}$ in Equation (\ref{eqn:hat_eps}) makes a new adversarial example $X_{adv}^{1}$ that ranges ($X_{adv}$ $+$ $\hat{\epsilon}_{0}$, $X_{adv}$ $-$ $\hat{\epsilon}_{0}$) where $X_{adv}$ for the lower boundary is $X$ $-$ $\epsilon$, $X_{adv}$ for the upper limit is $X$ $+$ $\epsilon$ and $\hat{\epsilon}_{0}$ is $\hat{\epsilon}$ of Equation (\ref{eqn:hat_eps}).
In order to find the upper and lower boundaries for $X_{adv}^{2}$ which is mitigated from $X_{adv}^{1}$, we need to estimate the perturbation $\hat{\epsilon}_{1}$ at both ends of the range for $X_{adv}^{1}$.
\begin{align}
\forall X_{adv} &- \hat{\epsilon}_{0}, \nonumber \\
\hat{\epsilon}_{1} &= (X_{adv} - \hat{\epsilon}_{0}) - W_{avg}*(X_{adv} - \hat{\epsilon}_{0}) \nonumber \\
									 &= (X + \epsilon - \hat{\epsilon}_{0}) - W_{avg}*(X + \epsilon - \hat{\epsilon}_{0}) \nonumber \\
									 &\approx \epsilon - \hat{\epsilon}_{0} - W_{avg}*(\epsilon - \hat{\epsilon}_{0}) \;\; (\because X \approx W_{avg}*X), \nonumber \\
\nonumber \\
\forall X_{adv} &+ \hat{\epsilon}_{0}, \nonumber \\
\hat{\epsilon}_{1} &= W_{avg} * (X_{adv} + \hat{\epsilon}_{0}) - (X_{adv} + \hat{\epsilon}_{0}) \nonumber \\
									 &= W_{avg} * (X - \epsilon + \hat{\epsilon}_{0}) - (X - \epsilon + \hat{\epsilon}_{0}) \nonumber \\
									 &\approx \epsilon - \hat{\epsilon}_{0} - W_{avg}*(\epsilon - \hat{\epsilon}_{0}) \;\; (\because X \approx W_{avg}*X) \nonumber 
\end{align}
where $\hat{\epsilon}_{0}$ $=$ $\epsilon$ $-$ $W_{avg}*\epsilon$ such that 
\begin{align}
\hat{\epsilon}_{1} &= W_{avg} * (\epsilon - W_{avg}*\epsilon) \nonumber \\
								   &= W_{avg} * \hat{\epsilon}_{0}. \nonumber 
\end{align}
Since it is more probable that $W_{avg}*|\hat{\epsilon}_{0}|$ $<$ $|\hat{\epsilon}_{0}|$, $|\hat{\epsilon}_{1}|$ $<$ $|\hat{\epsilon}_{0}|$.
Thus, $|\hat{\epsilon}_{1}|$ $<$ $|\hat{\epsilon}_{0}|$ $<$ $|\epsilon|$ since $|\hat{\epsilon}_{0}|$ $<$ $|\epsilon|$.
Figure \ref{fig:multilevel_mitigation_illustration} illustrates the multi-level mitigation that estimates $\hat{\epsilon}_{1}$.

In Figure \ref{fig:multilevel_mitigation_illustration}, the grey line is the boundary condition for $X_{adv}$ (corresponding to the solid black line of Figure \ref{fig:mitigation_with_hat_of_eps}), the solid black line denotes the range of $X_{adv}^1$ and the green line represents the upper and lower boundaries that $X_{adv}^2$ can reach.
As the level of mitigation goes deep, the accumulated sum of the estimated perturbations grows even though the estimated perturbation per each level decreases. Therefore, we should control $\hat{\epsilon}$ to satisfy the following.
\begin{align}
\sum_{n=0}^{p} |\hat{\epsilon}_{n}| &< |\epsilon| \;\;,\;\;  |\hat{\epsilon}_{p}| < \cdots < |\hat{\epsilon}_{1}| < |\hat{\epsilon}_{0}| < |\epsilon|  
\label{eqn:eps_cond}
\end{align}
where $p$ is the level of mitigation. However, in Equation (\ref{eqn:eps_cond}), $\epsilon$ cannot be directly handled as a single term because it is hidden in $X_{adv}$.
In order to control $\hat{\epsilon}$, Equation (\ref{eqn:eps_cond}) should be rephrased as below.
\begin{align}
X + \epsilon - \hat{\epsilon}_{0} - \hat{\epsilon}_{1} \cdots -\hat{\epsilon}_{p-1} -\hat{\epsilon}_{p} &> X-\epsilon, \nonumber \\
X - \epsilon + \hat{\epsilon}_{0} + \hat{\epsilon}_{1} \cdots +\hat{\epsilon}_{p-1} +\hat{\epsilon}_{p} &< X+\epsilon.
\label{eqn:hat_eps_cond}
\end{align}
In case that adjacent samples are very different to each other, $X$ $\not\approx$ $W_{avg}*X$ such that the estimated perturbation at the $p$-th step of the multi-level mitigation, $\hat{\epsilon}_{p}$ can be too large to satisfy Equation (\ref{eqn:hat_eps_cond}).
Since usually $X$  $>$ $\epsilon$, the large difference that breaks the relation of Equation (\ref{eqn:hat_eps_cond}) may come from the difference between the benign parts (i.e. $X$ of $X_{adv}$).
To relax the impact of the large differences on the estimated perturbations, we normalize the estimated perturbation at the $p$-th step of the multi-level mitigation as below.
\begin{align}
\hat{\epsilon}_{p} &= E[|X_{adv}^{p} - (W_{avg}*X_{adv}^{p})|] 
\label{eqn:hat_eps_normal}
\end{align}
where $E[|\cdot|]$ indicates both $E[X_{adv}^{p} - (W_{avg}*X_{adv}^{p})]$ (to be subtracted from $X_{adv}^{p-1}$) and $E[(W_{avg}*X_{adv}^{p})-X_{adv}^{p}]$ (to be added into $X_{adv}^{p-1}$).
This normalization also prevents $\hat{\epsilon}_{p}$ from working as a serious perturbation for $X_{adv}$ having $\epsilon$ $=$ $0$ ($\because$ the mitigation scheme should work for the benign example).

\begin{figure}[htb]
\begin{subfigure}[htb]{0.4\textwidth}
	\centering
  \includegraphics[width=\textwidth]{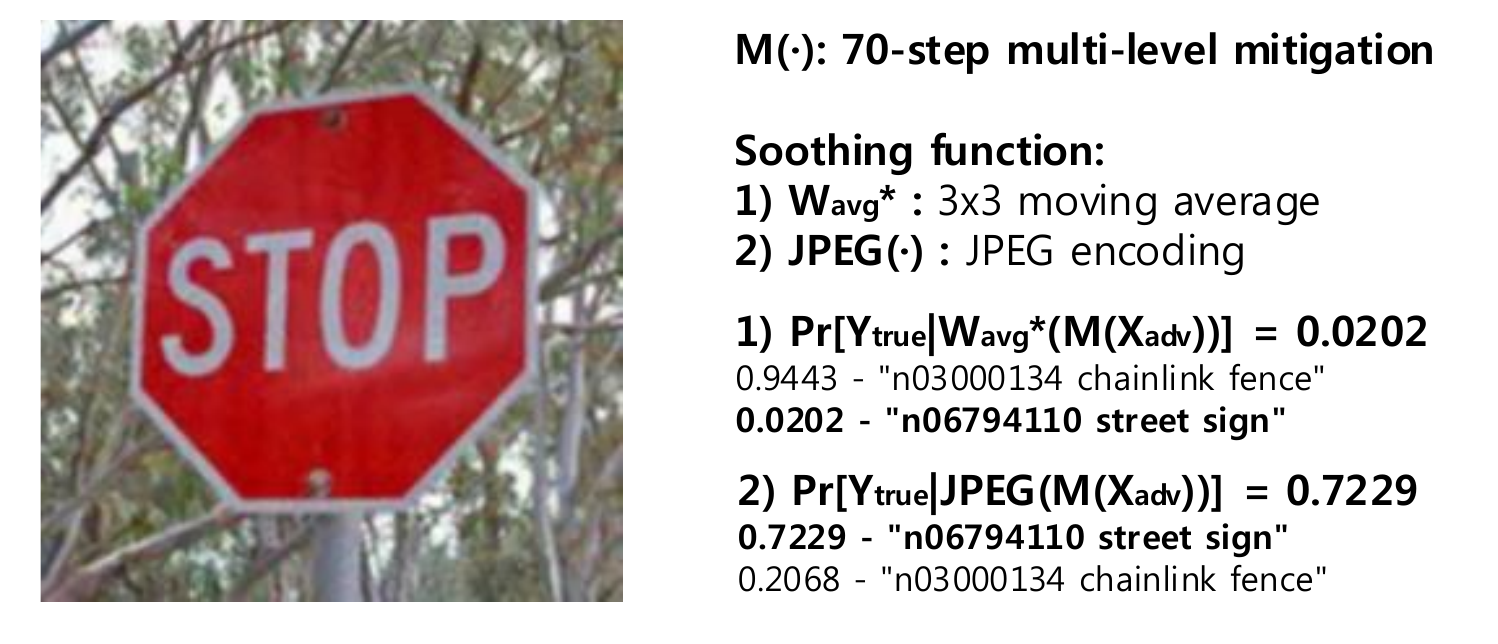}
  \caption{70-step multi-level mitigation gets higher prediction accuracy than the single-level mitigation of Figure \ref{fig:eps32_single}.}
  \label{fig:multi_with_ma}
\end{subfigure}
\begin{subfigure}[htb]{0.4\textwidth}
	\centering
  \includegraphics[width=\textwidth]{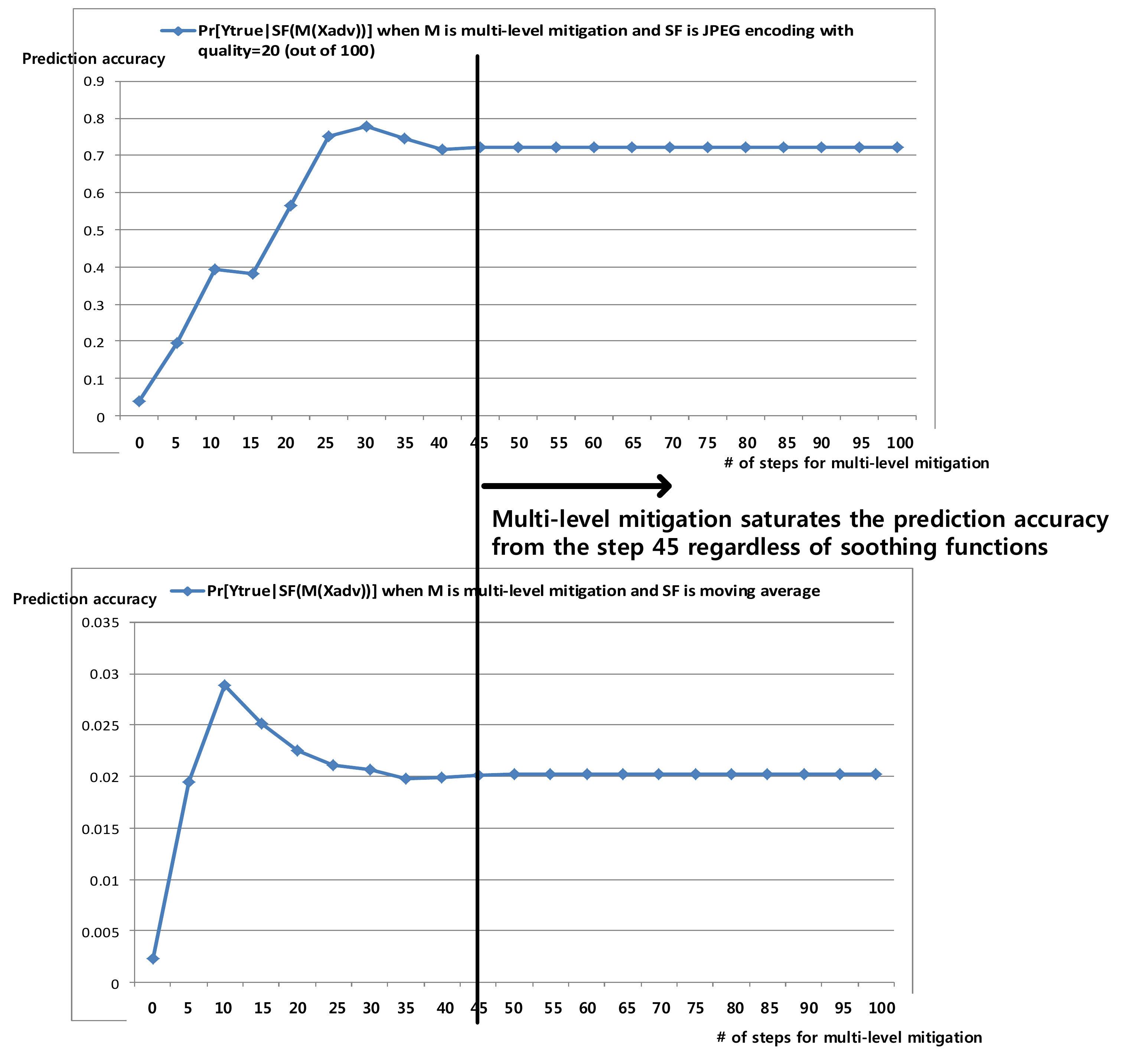}
  \caption{The boundary condition, Equation (\ref{eqn:hat_eps_bound}) for multi-level mitigation guarantees a certain level of the enhanced prediction regardless of soothing functions.}
  \label{fig:multi_with_boundary}
\end{subfigure}
\caption{The proposed multi-level mitigation enhances the prediction accuracy of Figure \ref{fig:eps32_single} in a controlled manner. }
\label{fig:multilevel_mitigation_demo}
\end{figure}

In Equation(\ref{eqn:hat_eps_cond}), the boundary condition for both decreasing perturbation $X_{adv}^{p-1}$ $-$ $\hat{\epsilon}$ and increasing perturbation $X_{adv}^{p-1}$ $+$ $\hat{\epsilon}$ better be replaced with the value closer to $X$.
As discussed earlier with Equation (\ref{eqn:moving_eps}), the probability for $W_{avg}*|\epsilon|$ $<$ $|\epsilon|$ is much larger than $W_{avg}*|\epsilon|$ $=$ $|\epsilon|$ such that it is more probable that $W_{avg}*X_{adv}$ is closer to $X$ than $X$ $\pm$ $\epsilon$ when $W_{avg}*X$ $\approx$ $X$.
That is,
\begin{align}
\forall X_{adv} &= X + \epsilon, \nonumber \\
W_{avg}*X_{adv} &= W_{avg} * ( X + \epsilon ) \nonumber \\
							  &= X + W_{avg} * \epsilon \;\; (\because W_{avg} * X \approx X) \nonumber \\
								&< X + \epsilon \;\; (\because W_{avg}*|\epsilon| < |\epsilon|), \nonumber \\
\nonumber \\
\forall X_{adv} &= X - \epsilon, \nonumber \\
W_{avg}*X_{adv} &= W_{avg} * ( X - \epsilon ) \nonumber \\
							  &= X - W_{avg} * \epsilon \;\; (\because W_{avg} * X \approx X) \nonumber \\
								&> X - \epsilon \;\; (\because W_{avg}*|\epsilon| < |\epsilon|). \nonumber 
\end{align}
Thus, both  $X$ $-$ $\epsilon$ and $X$ $+$ $\epsilon$ of Equation (\ref{eqn:hat_eps_cond}) can be replaced with $W_{avg}*X_{adv}$ as following.
\begin{align}
X_{adv}^{p-1} - \hat{\epsilon}_{p} &> W_{avg}*X_{adv},  \nonumber \\
X_{adv}^{p-1} + \hat{\epsilon}_{p} &< W_{avg}*X_{adv}.  
\label{eqn:hat_eps_bound}
\end{align}
Equation (\ref{eqn:hat_eps_bound}) guarantees that the proposed multi-level mitigation gets $X_{adv}$ closer to $X$ with the smaller (i.e. mitigated) perturbation having the same polarity with the original (i.e. unmitigated) perturbation if $X$ $\approx$ $W_{avg}*X$.
That is, the term of decreasing $\epsilon$, $X_{adv}^{p-1}$  $-$ $\hat{\epsilon}_{p}$ does not go below $W_{avg} * X_{adv}$ and the term of increasing $\epsilon$, $X_{adv}^{p-1}$ $+$ $\hat{\epsilon}_{p}$ does not go beyond $W_{avg} * X_{adv}$.
To this end, $W_{avg}*X_{adv}$ can be used as the decision boundary that determines the maximum mitigation step where the prediction accuracy does not change any longer. Figure \ref{fig:multilevel_mitigation_demo} shows that the 70-step multi-level mitigation satisfying Equation (\ref{eqn:hat_eps_bound}) is applied to the adversarial example having $\epsilon$ $=$ 32 of Figure \ref{fig:single_mitigation}.

In Figure \ref{fig:multilevel_mitigation_demo}, even though the prediction accuracies of Figure \ref{fig:multi_with_ma} are different according to soothing filters, the mitigation step where their prediction accuracies do not change any longer is same as shown in Figure \ref{fig:multi_with_boundary}.
This means that the boundary conditions of Equation (\ref{eqn:hat_eps_bound}) guarantee a certain level of prediction accuracy if the number of mitigation steps are large enough like the 70-step multi-level mitigation of Figure \ref{fig:multilevel_mitigation_demo}.
Thanks to Equation (\ref{eqn:hat_eps_bound}), we can build the algorithmic state machine having the point where the multi-level mitigation stops at.

\subsection{Algorithm}
Algorithm \ref{alg:mitigating_perturbation} summarizes the proposed method of mitigating adversarial perturbations.
\begin{algorithm}
   \caption{Mitigating adversarial perturbations}
   \label{alg:mitigating_perturbation}
\begin{footnotesize}
\begin{algorithmic}[1]
	 \State //$X_{adv}$ is the adversarial example.
   \State //$\hat{\epsilon}$ is the estimated perturbation.
   \State //$W_{avg}$ is $N \times N$ moving average kernel.
   \State //$X_{adv}^{j}$ is the adversarial example at the j-th step mitigation
   \State //$\hat{\epsilon}_{j}$ is the estimated perturbation at the j-step mitigation
   \State //$W_{avg}*X_{adv}$ is the local expectation that $W_{avg}$ covers.
   \State //$k$ is the size of an array having CNN inference results from $X_{adv}^{j-(k-1)}$ to $X_{adv}^{j}$
   \State //$SF$ is the soothing filter like JPEG encoding
   \State //CNN is a convolutional neural network 
	 \State
   \State {\bfseries Input:} $X_{adv}$, $W_{avg}$, k, SF, CNN
	 \State //initial step j is set to zero
	 \State j = 0;
	 \State //initialize the variable that counts the number of consecutive k-steps 
	 \State //whose prediction accuracies are same
	 \State equal\_count = 0;
	 \Repeat 
	 \If { $X_{adv}^{j}$ $\geq$ ($W_{avg} * X_{adv}^{j}$) }
	 \State //Find $\hat{\epsilon}_{j}$ to be subtracted from $X_{adv}^{j-1}$	
   \State $\hat{\epsilon}_{j}$ $=$ $X_{adv}^{j}$ $-$ ($W_{avg} * X_{adv}^{j}$);
	 \State //Mitigate the perturbation with $\hat{\epsilon}_{j}$ iff $\hat{\epsilon}_{j}$ $\leq$ $\hat{\epsilon}_{j-1}$
	 \If { (j $>$ 0 ) \&\& ( $\hat{\epsilon}_{j}$ $\leq$ $\hat{\epsilon}_{j-1}$ ) }
	 \State //Subtract $\hat{\epsilon}_{j}$ from $X_{adv}^{j-1}$
	 \State //if and only if this subtraction meets Equation (\ref{eqn:hat_eps_bound})
   \If {($X_{adv}^{j-1}$ $-$ $\hat{\epsilon}_{j}$) $>$ $W_{avg}*X_{adv}$}
	 \State $X_{adv}^{j}$ $=$ $X_{adv}^{j-1}$ $-$ $\hat{\epsilon}_{j}$;
	 \Else
	 \State $X_{adv}^{j}$ $=$ $X_{adv}^{j-1}$;
	 \EndIf
	 \EndIf
   \Else  
	 \State //Find $\hat{\epsilon}_{j}$ to be added to $X_{adv}^{j-1}$
	 \State $\hat{\epsilon}_{j}$ $=$ ($W_{avg} * X_{adv}^{j}$) $-$ $X_{adv}^{j}$;
   \State //Mitigate the perturbation with $\hat{\epsilon}_{j}$ iff $\hat{\epsilon}_{j}$ $\leq$ $\hat{\epsilon}_{j-1}$
	 \If {(j $>$ 0 ) \&\& ( $\hat{\epsilon}_{j}$ $\leq$ $\hat{\epsilon}_{j-1}$ )}
	 \State //Add $\hat{\epsilon}_{j}$ to $X_{adv}^{j-1}$
   \State //if and only if this addition satisfies Equation (\ref{eqn:hat_eps_bound})
	 \If { ($X_{adv}^{j-1}$ $+$ $\hat{\epsilon}_{j}$) $<$ $W_{avg}*X_{adv}$ }
	 \State $X_{adv}^{j}$ $=$ $X_{adv}^{j-1}$ $+$ $\hat{\epsilon}_{j}$;
	 \Else
	 \State $X_{adv}^{j}$ $=$ $X_{adv}^{j-1}$;
   \EndIf
   \EndIf
   \EndIf
   \State //Stop the mitigation if the results of consecutive k-steps are same
   \State //Pop-and-Push to store new result
   \For{ $i=0$ {\bfseries to} $k-1$}
	 \State result[k-i] = result[k-(i+1)];
	 \EndFor
   \State result[0] = CNN(SF($X_{adv}^{j}$);
   \State //Check if the results of consecutive k-steps are same or not
   \For{ $n=0$ {\bfseries to} $k-2$}
	 \If{result[n] != result[n+1]}
	 \State break;
	 \EndIf
	 \EndFor
	 \State equal\_count = n;
	 \Until{ equal\_count is $k-2$}
\end{algorithmic}
\end{footnotesize}
\end{algorithm}

Among the inputs of Algorithm \ref{alg:mitigating_perturbation}, $X_{adv}$ and $W_{avg}$ are used to estimate $\hat{\epsilon}_{j}$ and other parameters are related to the condition to stop the multi-level mitigation.
As demonstrated in the previous sections, the proposed mitigation scheme has no dependency on the soothing filter and neural network architecture.
However, the soothing filter and a CNN inference are required to check if the prediction accuracy gets saturated or not.
Also, there should be an array having the size of k (which stores the previous prediction results) as the input parameter of the algorithm in order to see if the prediction accuracy of the current mitigation step is saturated or not.

\section{Evaluation of proposed method} \label{sec:evaluation}
For the evaluation of the proposed mitigation scheme described in Algorithm \ref{alg:mitigating_perturbation}, 
$\epsilon$ is  crafted by FGSM (Fast Gradient Sign Method) attacks~\cite{alexey} on ResNet-50~\cite{resnet} trained with ImageNet~\cite{imagenet}.
Thus, the input parameter "CNN" of Algorithm \ref{alg:mitigating_perturbation} is ResNet-50. 
Also, both JPEG encoding and $3 \times 3$ moving average filter for the variable "SF". 
We repeat the mitigation until the prediction accuracy gets into a saturated point such that we do not specify the parameter $k$ for this evaluation.
The core part (i.e. estimating perturbations and checking the boundary condition for multi-level mitigation) of Algorithm \ref{alg:mitigating_perturbation} are written in "convert" script of ImageMagick~\cite{imagemagick}. 
The script to find $\hat{\epsilon}$ with 3 $\times$ 3 $W_{avg}$ is found in Appendix \ref{sec:appendix_code} 
\begin{footnotesize}
\footnote{Other scripts and adversarial examples used for this evaluation are avaiable at \url{https://github.com/stonylinux/mitigating_large_adversarial_perturbations_on_X-MAS}} 
\end{footnotesize}
.

\begin{figure}[htb]
\begin{subfigure}[htb]{0.45\textwidth}
	\centering
  \includegraphics[width=\textwidth]{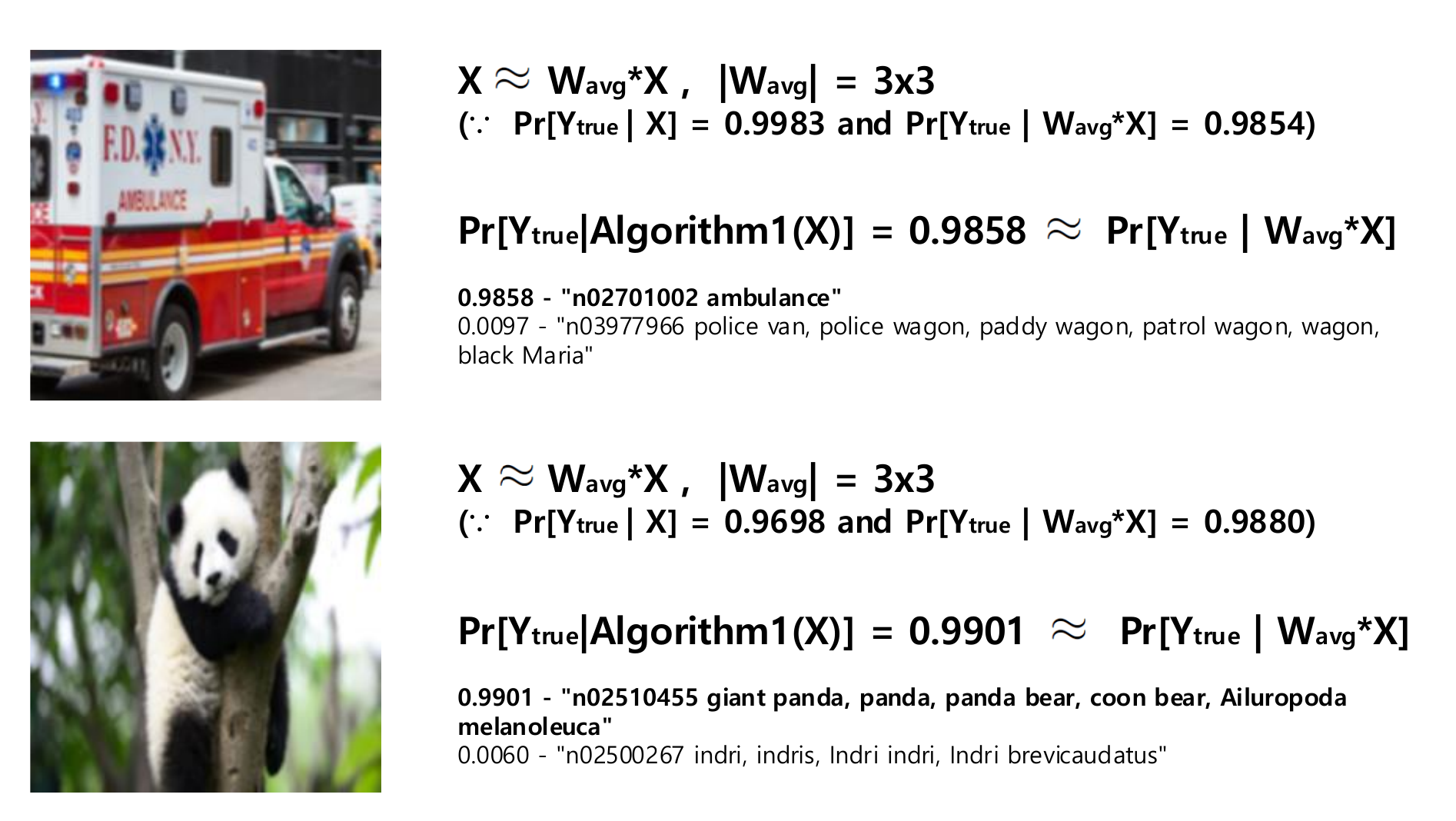}
  \caption{$Pr[Y_{true}|Algorithm1(X)]$ $\approx$ $Pr[Y_{true}|X]$ when $Pr[Y_{true}|X]$ $\approx$  $Pr[Y_{true}|W_{avg}*X]$.}
  \label{fig:eps0_miti_ambulance_panda}
\end{subfigure}
\begin{subfigure}[htb]{0.45\textwidth}
	\centering
  \includegraphics[width=\textwidth]{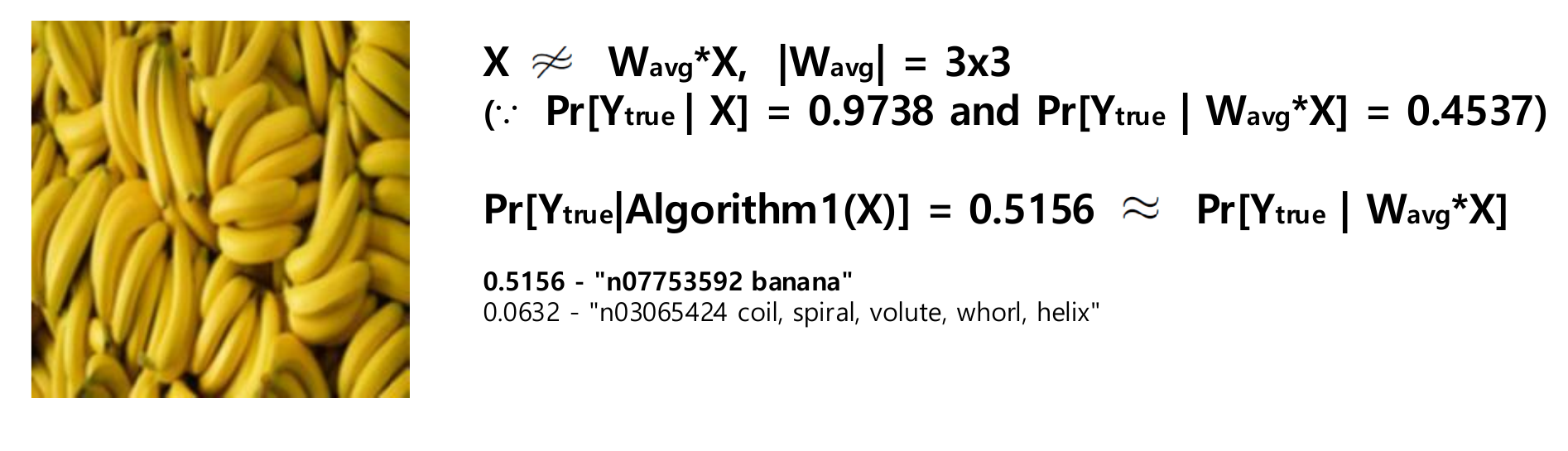}
  \caption{$Pr[Y_{true}|Algorithm1(X)]$ $\not\approx$ $Pr[Y_{true}|X]$ but $Pr[Y_{true}|Algorithm1(X)]$ $\approx$ $Pr[Y_{true}|W_{avg}*X]$ when $Pr[Y_{true}|X]$ $\not\approx$  $Pr[Y_{true}|W_{avg}*X]$.}
  \label{fig:eps0_miti_banana}
\end{subfigure}
\caption{Algorithm \ref{alg:mitigating_perturbation} makes $Pr[Y_{true}|Algorithm1(X)]$ $\approx$ $Pr[Y_{true}|W_{avg}*X]$.}
\label{fig:eps0_mitigated}
\end{figure}

First of all, the proposed mitigation scheme should work for the case where $\epsilon$ $=$ 0 since it does not know if its input image has any adversarial perturbation or not. Algorithm \ref{alg:mitigating_perturbation} keeps mitigating the perturbation unless it crosses over the boundary condition of $W_{avg}*X_{adv}$. When $X_{adv}$ has no perturbation, the boundary condition becomes $W_{avg} * X$. 
Also, the estimated perturbation $\hat{\epsilon}$ comes from the difference between $X$ and $W_{avg}*X$.
Since the proposed scheme is valid for the case $X$ $\approx$ $W_{avg}*X$, $\hat{\epsilon}$ for $\epsilon$ does not make the prediction accuracy $Pr[Y_{true}|Algorithm1(X)]$ far different from $Pr[Y_{true}|W*X]$ (i.e. $Pr[Y_{true}|Algorithm1(X)]$ $\approx$ $Pr[Y_{true}|W*X]$) as shown in Figure \ref{fig:eps0_mitigated}.

In Figure \ref{fig:eps0_miti_ambulance_panda}, $Pr[Y_{true}|Algorithm1(X)]$ is slightly better than $Pr[Y_{true}|W_{avg}*X]$. This means that $X$ is purely mitigated (i.e. perturbed) by the difference of benign parts among adjacent samples but the mitigation does not work as a serious perturbation since the difference (of benign parts among adjacent samples) cannot make $X$ cross over the boundary condition $W_{avg}*X$ according to Equation (\ref{eqn:hat_eps_bound}).
When $X$ $\not\approx$ $W_{avg}*X$ 
in Figure \ref{fig:eps0_miti_banana}, Algorithm \ref{alg:mitigating_perturbation} keeps mitigating the perturbation (i.e. adding/subtracting the difference between $X$ and $W_{avg}*X$ to/from $X$)
 until $X_{adv}^{p}$ ($X_{adv}$ in p-step mitigation, actually $X^{p}$ for the case $\epsilon$ $=$ $0$) has the minimum distance from $W_{avg}*X$.
As the result, $Pr[Y_{true}|Algorithm1(X)]$ $\approx$ $Pr[Y_{true}|W_{avg}*X]$ even though $Pr[Y_{true}|Algorithm1(X)]$ $\not\approx$ $Pr[Y_{true}|X]$.

\begin{figure}[htb]
\begin{subfigure}[htb]{0.45\textwidth}
	\centering
  \includegraphics[width=\textwidth]{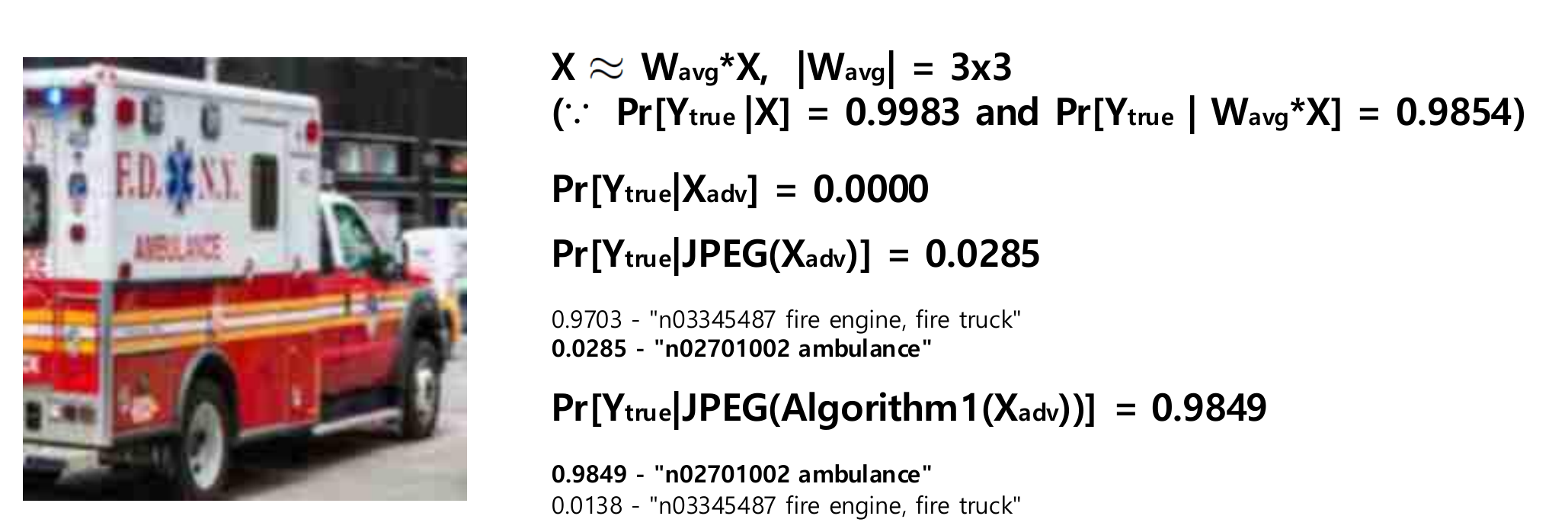}
  \caption{Algorithm \ref{alg:mitigating_perturbation} increases the prediction accuracy from 0.0285 to 0.9849 when the basic iterative FGSM attack with $\epsilon$ $=$ $32$ is applied.}
  \label{fig:ambulance_miti}
\end{subfigure}
\begin{subfigure}[htb]{0.45\textwidth}
	\centering
  \includegraphics[width=\textwidth]{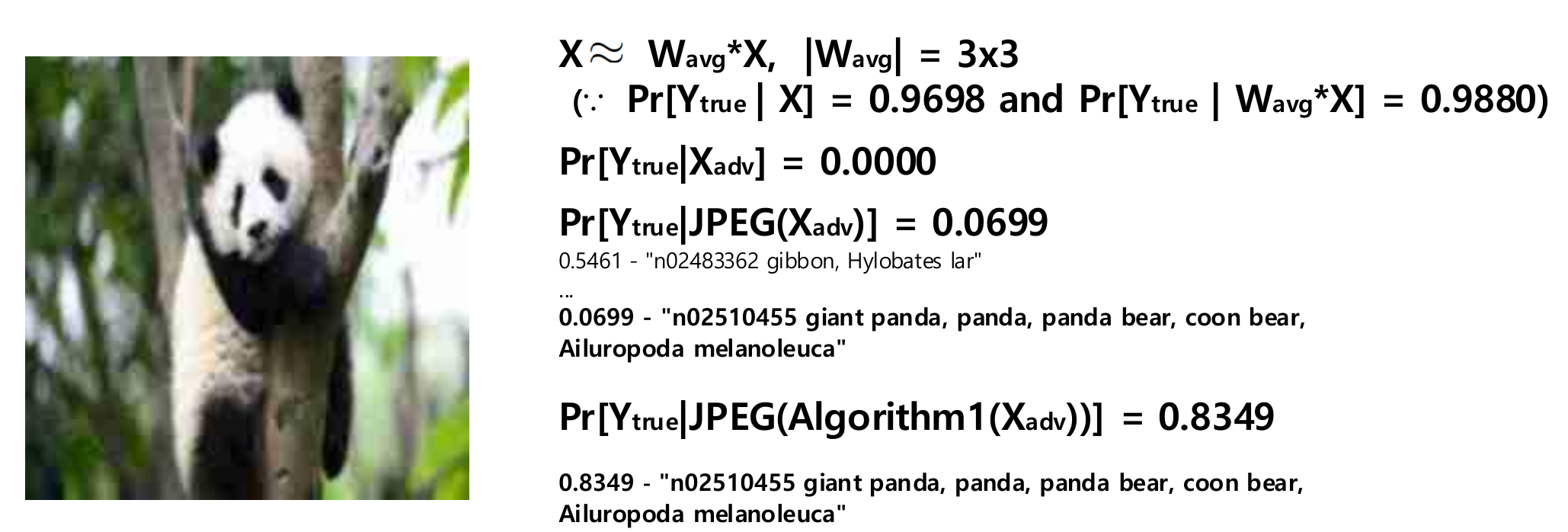}
  \caption{Algorithm \ref{alg:mitigating_perturbation} improves the prediction accuracy from 0.0699 to 0.8349 when the basic iterative FGSM attack with $\epsilon$ $=$ $32$ is applied.}
  \label{fig:panda_miti}
\end{subfigure}
\caption{The proposed mitigation contributes to the enhancement of prediction accuracy much more than JPEG encoding does.}
\label{fig:well_mitigated}
\end{figure}

Most moving average outcomes for the input images of a CNN are predicted as the same label with their original images (i.e. $Pr[Y_{true}|X]$ $\approx$ $Pr[Y_{true}|W_{avg}*X]$) because the moving average convolution of the input image works as the low-pass filter like human eyes.
Figure \ref{fig:well_mitigated} shows that Algorithm \ref{alg:mitigating_perturbation} well estimates $\hat{\epsilon}_{j}$ when $X$ $\approx$ $W_{avg}*X$. 

\begin{figure}[htb]
\begin{subfigure}[htb]{0.45\textwidth}
	\centering
  \includegraphics[width=\textwidth]{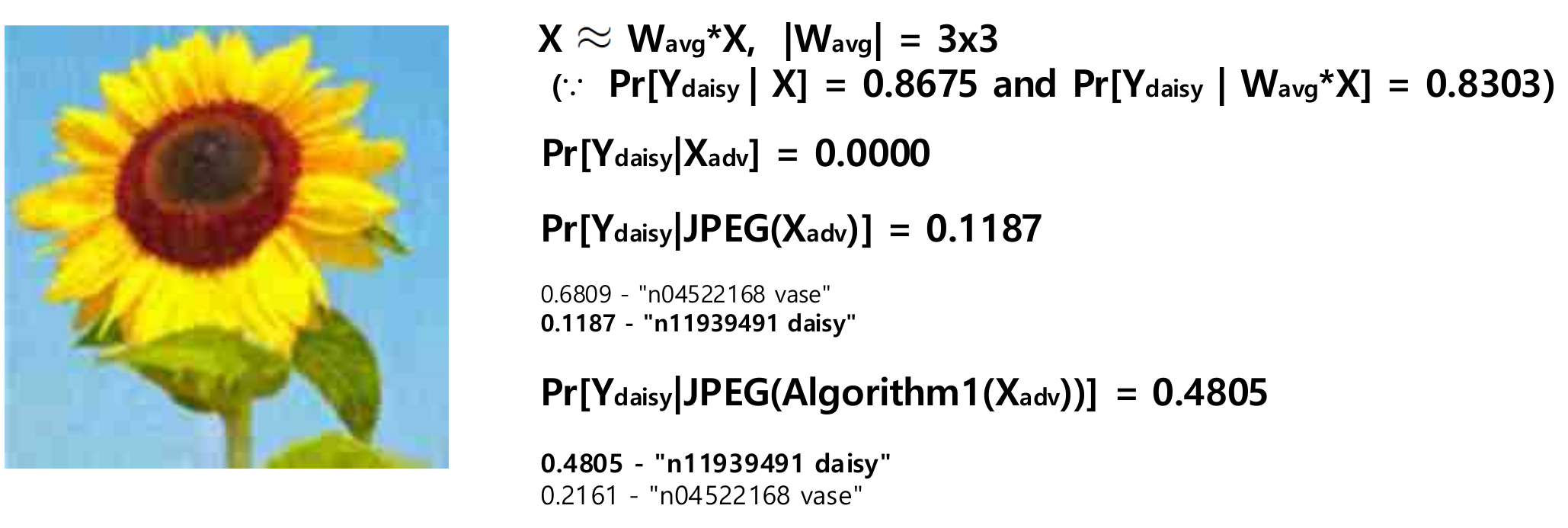}
  \caption{For the CNN recognizing sunflower as daisy, $\hat{\epsilon}$ can be found if $Pr[Y_{daisy}|X]$ $\approx$ $Pr[Y_{daisy}|W_{avg}*X]$ ($X_{adv}$ is generated by the basic iterative FGSM attack with $\epsilon$ $=$ 32).}
  \label{fig:sunflower_miti}
\end{subfigure}
\begin{subfigure}[htb]{0.45\textwidth}
	\centering
  \includegraphics[width=\textwidth]{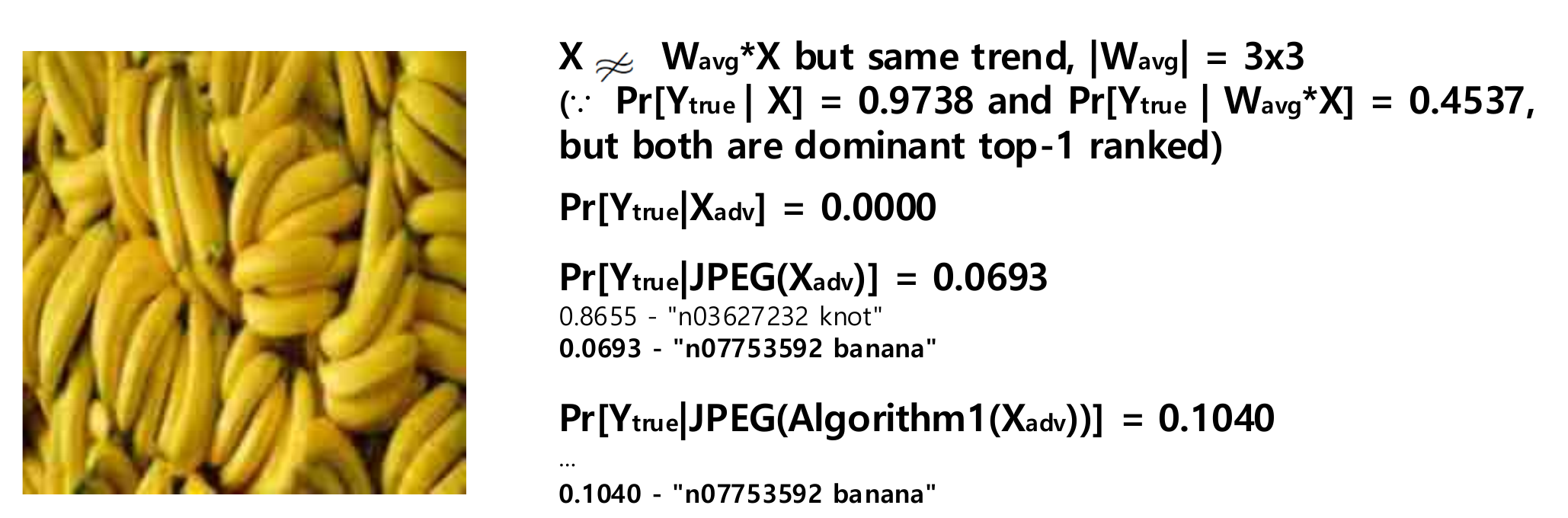}
  \caption{The proposed mitigation might not increase the prediction accuracy very much if $Pr[Y_{true}|X]$ $\not\approx$ $Pr[Y_{true}|W_{avg}*X]$ ($X_{adv}$ is generated by the basic iterative FGSM attack with $\epsilon$ $=$ 32).}
  \label{fig:banana_miti}
\end{subfigure}
\caption{The proposed mitigation requires that $Pr[Y|X]$ $\approx$ $Pr[Y|W_{avg}*X]$ rather than that the ranks of prediction accuracies are close.}
\label{fig:soso_mitigated}
\end{figure}

In Figure \ref{fig:well_mitigated}, the rank of prediction accuracy should be carefully handled. When JPEG encoding is used as the soothing filter without the mitigation steps (i.e. for the case of $Pr[Y_{true}|JPEG(X_{adv})]$), it achieves top-2 accuracy for Figure \ref{fig:ambulance_miti} as well as it gets top-3 accuracy for Figure \ref{fig:panda_miti}. 
However, both cases work well as the adversarial examples having dominant prediction accuracies. 
Especially for $Pr[Y_{true}|JPEG(X_{adv})]$ of Figure \ref{fig:ambulance_miti}, no one can say it is okay to reach the top-2 accuracy with the probability of 0.0285 because the wrong recognition can be fatal to a human life (i.e. fire truck cannot replace the ambulance).
Thus, when we talk about the prediction accuracy, the number in probability can be much more important than the rank of the accuracy.
Figure \ref{fig:soso_mitigated} shows the cases that the number in probability plays an important role in the proposed mitigation.

\begin{figure}[htb]
  \centering
  \includegraphics[width=1.0\linewidth]{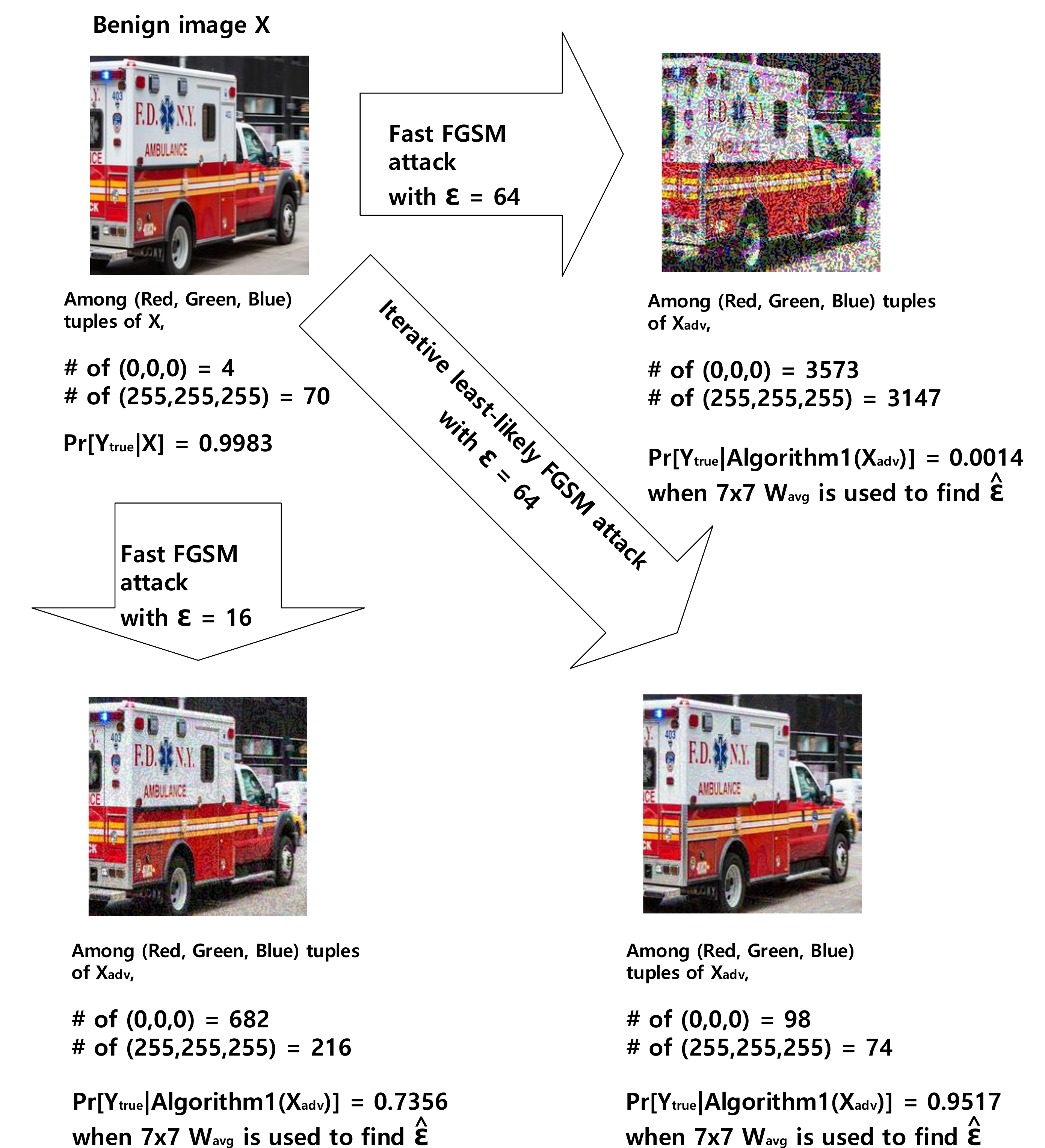}
  \caption{Algorithm \ref{alg:mitigating_perturbation} works better for the adversarial example which has the smaller number of out-of-bound samples.}
  \label{fig:other_FGSMs}
\end{figure}

Figure \ref{fig:soso_mitigated} shows the cases that $Pr[Y_{true}|X]$ $\not\approx$ $Pr[Y_{true}|W_{avg}*X]$. Figure \ref{fig:sunflower_miti} does not have $Y_{true}$ ($\because$ ImageNet dataset does not have the label corresponding to sunflower) and Figure \ref{fig:banana_miti} has far different numbers for $Pr[Y_{true}|X]$ and $Pr[Y_{true}|W_{avg}*X]$.
In Figure \ref{fig:sunflower_miti}, $Pr[Y_{daisy}|X]$ becomes the ground truth because the CNN inference having no label for sunflower recognizes $X$ as daisy (i.e. $Y_{daisy}$).
Algorithm \ref{alg:mitigating_perturbation} increases the prediction accuracy for $Y_{daisy}$ from 0.1187 to 0.4805 because $Pr[Y_{daisy}|X]$ $\approx$ $Pr[Y_{daisy}|W_{avg}*X]$. 
However, in Figure \ref{fig:banana_miti},
even though both ranks of the prediction accuracies $Pr[Y_{true}|X]$ and $Pr[Y_{true}|W_{avg}*X]$ are dominant (i.e. top-1 rank),
the proposed mitigation does not improve the prediction accuracy very much because $Pr[Y_{true}|X]$ $\not\approx$ $Pr[Y_{true}|W_{avg}*X]$ (i.e. the difference between $X$ and $W_{avg}*X$ works as a perturbation).
Also, larger $|W_{avg}|$ is better for Algorithm \ref{alg:mitigating_perturbation} if the corresponding $W_{avg}$ satisfies $Pr[Y_{true}|X]$ $\approx$ $Pr[Y_{true}|W_{avg}*X]$ with a high probability. It is because $\hat{\epsilon}$ (i.e. the difference between $X_{adv}$ and $W_{avg}*X_{adv}$) can be better normalized with larger $|W_{avg}|$. The details of the evaluation related to $W_{avg}$ are in Appendix \ref{sec:appendix_alg}.
For the FGSM attack without clipping function ("fast" attack in~\cite{alexey}), the large perturbation can seriously distort an image by generating out-of-bound samples. That is, it can make the samples in the range (0, $\epsilon$) become zero as well as it can increase the samples within (255 $-$ $\epsilon$, 255) to have the maximum value, 255.
Then, Algorithm \ref{alg:mitigating_perturbation} would not well mitigate the perturbations because the size of estimated perturbation $|\hat{\epsilon}|$ can be reduced by the amount of $|\epsilon - X|$ for the sample $X$ in the range (0, $\epsilon$) or by the amount of $|X-(255-\epsilon)|$ for $X$ in the range (255 $-$ $\epsilon$, 255).
Figure \ref{fig:other_FGSMs} shows how the performance of Algorithm \ref{alg:mitigating_perturbation} varies according to the numbers of out-of-bound samples.

In Figure \ref{fig:other_FGSMs}, fast FGSM attack is used to generate two distinct numbers of out-of-bound samples, which show the impact of the out-of-bound samples on the performance of Algorithm \ref{alg:mitigating_perturbation}. 
The statistical numbers of (Red, Green, Blue) tuples come from the ImageMagick convert script with the option "$-$format $\%$c histogram:info: ".
In fast FGSM attacks, $\epsilon$ is the knob to control the number of out-of-bound samples. 
The number of out-of-bound samples increases when $\epsilon$ goes up. Also, the number of out-of-bound samples decreases as $\epsilon$ goes down.
In Figure \ref{fig:other_FGSMs}, the adversarial example generated by the fast FGSM attack with $\epsilon$ $=$ $16$ has the smaller number of out-of-bound samples than the adversarial example that the fast FGSM attack with $\epsilon$ $=$ $64$ is applied to (i.e. 682 vs 3573 for (0,0,0) tuples and 216 vs 3147 for (255,255,255) tuples). 
Larger number of out-of-bound samples makes more perceptible perturbations.
Algorithm \ref{alg:mitigating_perturbation} shows the better performance (i.e. prediction accuracy represented as $Pr[Y_{true}|Algorithm1(X_{adv})]$) when the number of out-of-bound samples is small (i.e. when the perturbation is imperceptible).
The number of out-of-bound samples can be very different according to the attack methods.
For example,
the adversarial example crafted by the iterative least-likely FGSM attack with $\epsilon$ $=$ $64$ has the smaller number of out-of-bound samples than the adversarial example generated by the fast FGSM attack with $\epsilon$ $=$ $16$.
It is because the iterative least-likely FGSM attack has the clipping function that reduces the number of out-of-bound samples.

\section{Conclusion} \label{sec:conclusion}
In this paper, we propose the scheme that mitigates the perturbations on an adversarial example through the estimation of the perturbations on X-MAS (X minus Moving Average Samples i.e. when $X$ $\approx$ $W_{avg}*X$).
For large perturbations (i.e. $\epsilon$ $>$ $16$), the scheme is further developed to run the multi-step mitigation that has $W_{avg}*X_{adv}$ as the boundary condition to prevent the p-step mitigated adversarial example $X_{adv}^{p}$ from getting worse by crossing over the boundary.
The multi-level mitigation gets $X_{adv}$ closer to $X$ for the most adversarial examples whose benign part have the relation that $X$ $\approx$ $W_{avg}*X$. 
We evaluate the proposed algorithm with some representing examples that have the different relations between $X$ and $W_{avg}*X$. 
In the evaluation, our proposed scheme well mitigates the imperceptibly crafted large adversarial perturbations (through iterative FGSM attacks with a clipping function) such that it gets the adversarial examples with the mitigated perturbations have high prediction accuracies when $X$ $\approx$ $W_{avg}*X$.

\bibliographystyle{./icml2019}
\bibliography{mitigating_perturbation}


\appendix
\newpage
\twocolumn[
\section{The script for 100-step multi-level mitigation } \label{sec:appendix_code}
\begin{scriptsize}
\begin{lstlisting}
#!/bin/bash
#
# Version for 3x3 estimation
# 
# To mitigate the adversarial example "adversarial_example.png"
#
# ./mitigating_adversarial_with_3x3_estimation.sh adversarial_example
#
# Caution!!! All "convert" commands better be typed in one line
# because -fx content within ' ' does not support bash typed linking with "\"
# The reason we split convert command here is just for the better display
#

# Initial mitigation starts
#
# Find W_{avg}*X_{adv}
convert $1.png -fx '((p[-1,-1]+p[-1,0]+p[-1,1]+p[0,-1]+p[0,1]+p[1,-1]+p[1,0]+p[1,1])+p[0,0])/9' $1_ma0.png

# Find X_{adv} - W_{avg}*X_{adv} for X_{adv} > W_{avg}*X_{adv}
convert $1.png $1_ma0.png -fx 'u[0].p[0,0]-u[1].p[0,0]' $1_meps0.png

# Find W_{avg}*X_{adv} - X_{adv} for X_{adv} < W_{avg}*X_{adv}
convert $1_ma0.png $1.png -fx 'u[0].p[0,0]-u[1].p[0,0]' $1_peps0.png

# Check the upper and lower bound of the mitigated samples
#
# Following is just for displayed
convert $1.png $1_meps0.png $1_peps0.png -fx '( (u[1].p[0,0]> 0) &&  (u[2].p[0,0]== 0) && 
((u[0].p[0,0] - u[1].mean) > (u[0].minima)) )?u[0].p[0,0] - u[1].mean:( (u[1].p[0,0] == 0) &&  
(u[2].p[0,0]> 0) && ((u[0].p[0,0]+u[2].mean) < (u[0].maxima)) )?u[0].p[0,0]+u[2].mean:u[0].p[0,0]' 
$1_mitigated_step0.png


# Initial mitigation ends


# iterative mitigation starts 
for i in {0..99}
do
	convert $1_mitigated_step$i.png -fx '((p[-1,-1]+p[-1,0]+p[-1,1]+p[0,-1]+p[0,1]+p[1,-1]+p[1,0]+p[1,1])
	+p[0,0])/9' $1_ma$((i+1)).png

	convert $1_mitigated_step$i.png $1_ma$((i+1)).png -fx 'u[0].p[0,0]-u[1].p[0,0]' $1_meps$((i+1)).png

	convert $1_ma$((i+1)).png $1_mitigated_step$i.png -fx 'u[0].p[0,0]-u[1].p[0,0]' $1_peps$((i+1)).png

	# Check the upper and lower bound of the mitigated samples
	#
	# DO NOT mitigate X_{adv}^{p-1}  
	#
	# if $\hat{\epsilon}_{p} \geq \hat{\epsilon}_{p-1}$
	# or
	# if \hat{\epsilon}_{p} \geq \hat{\epsilon}_{p-1}
	# or
	# if X_{adv}^{p-1} - \hat{\epsilon}_{p} < W_{avg}*X_{adv} (for X_{adv}^{p-1} > W_{avg}*X_{adv})
	# or
	# if X_{adv}^{p-1} + \hat{\epsilon}_{p} > W_{avg}*X_{adv} (for X_{adv}^{p-1} < W_{avg}*X_{adv})
	#
	convert $1_mitigated_step$i.png $1_meps$((i+1)).png $1_peps$((i+1)).png $1_meps$i.png $1_peps$i.png 
	$1_ma0.png  -fx '( (u[1].p[0,0]> 0) &&  (u[2].p[0,0]== 0) && ((u[0].p[0,0] - u[1].mean) >= (u[5].p[0,0])) 
	&& (u[1].mean < u[3].mean) )?u[0].p[0,0] - u[1].mean:( (u[2].p[0,0]> 0) && (u[1].p[0,0] == 0) &&  
	((u[0].p[0,0]+u[2].mean) <= (u[5].p[0,0])) && (u[2].mean < u[4].mean) )?u[0].p[0,0]+u[2].mean:u[0].p[0,0]'
	$1_mitigated_step$((i+1)).png
done
# iterative mitigation ends

# Soothing filter part here!
# 3x3 Moving average as a soothing filter
convert $1_mitigated_step100.png  -fx '((p[-1,-1]+p[-1,0]+p[-1,1]+p[0,-1]+p[0,1]+p[1,-1]+p[1,0]+p[1,1])+p[0,0])/9' 
$1_mitigated_and_soothed_by_ma.png

# JPEG encoding with the quality 20 (out of 100)
convert $1_mitigated_step100.png -quality 20 $1_mitigated_and_soothed_by_JPEG.jpg

\end{lstlisting}
\end{scriptsize}
]

\newpage
\clearpage
\section{Evaluation of Algorithm \ref{alg:mitigating_perturbation} related to a moving-average kernel} \label{sec:appendix_alg}
\begin{figure}[htb]
\begin{subfigure}[htb]{0.5\textwidth}
  \includegraphics[width=\textwidth]{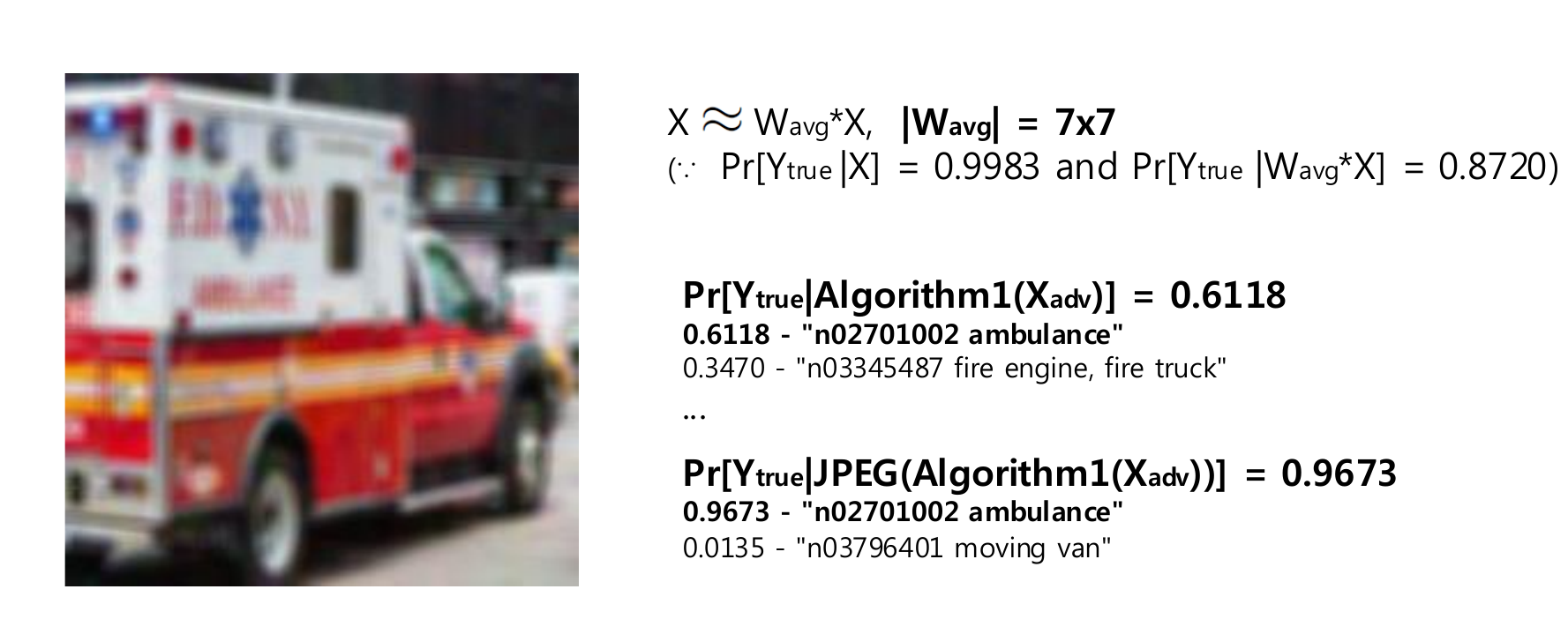}
  \caption{ $|W_{avg}|$ contributes more to the prediction accuracy when $|W_{avg}|$ is large and $Pr[Y_{true}|X]$ $\approx$ $Pr[Y_{true}|W_{avg}*X]$ ($X_{adv}$ is generated by the basic iterative FGSM attack with $\epsilon$ $=$ 64).}
  \label{fig:miti_ambulance_with_large_W}
\end{subfigure}
\begin{subfigure}[htb]{0.5\textwidth}
  \includegraphics[width=\textwidth]{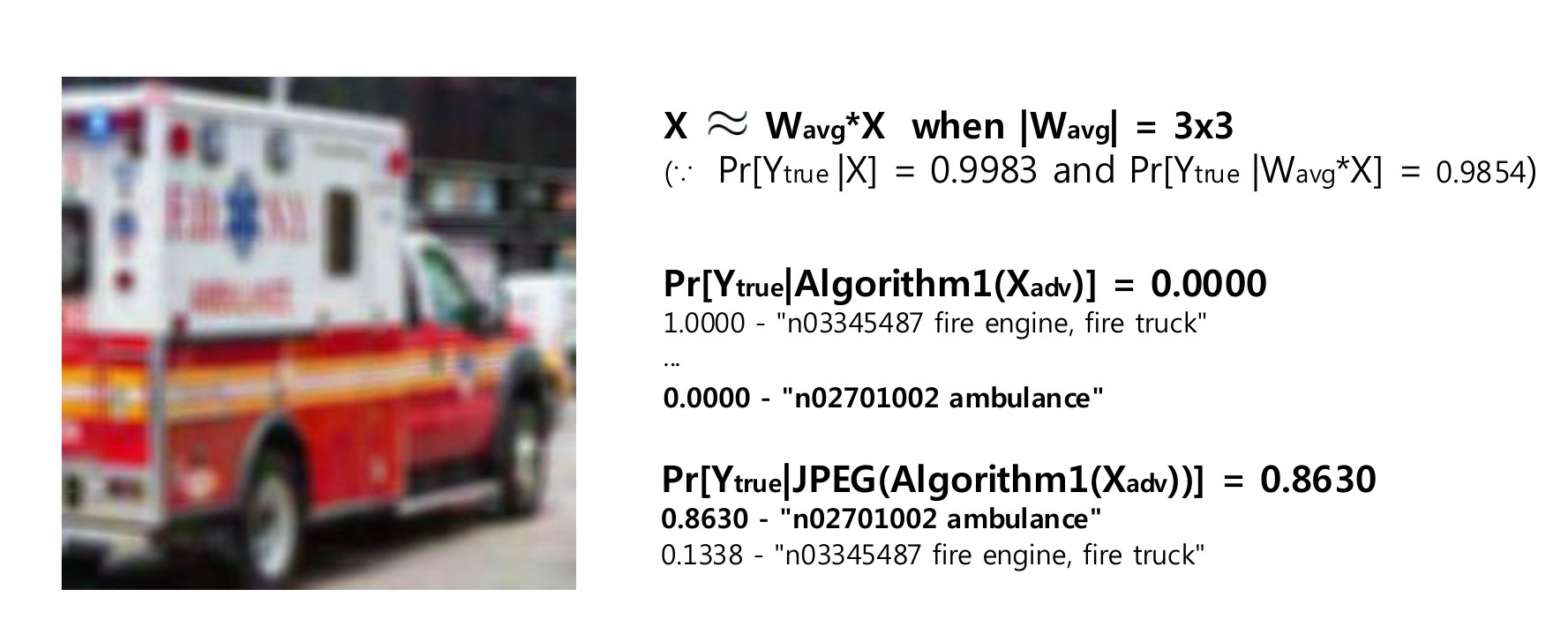}
  \caption{Soothing filter (i.e. JPEG) contributes more to the prediction accuracy when when $|W_{avg}|$ is small and $Pr[Y_{true}|X]$ $\approx$ $Pr[Y_{true}|W_{avg}*X]$ ($X_{adv}$ is generated by the basic iterative FGSM attack with $\epsilon$ $=$ 64).}
  \label{fig:miti_ambulance_with_small_W}
\end{subfigure}
\caption{Large $|W_{avg}|$ can make the proposed mitigation get a high prediction accuracy without soothing filter if $Pr[Y_{true}|X]$ $\approx$ $Pr[Y_{true}|W_{avg}*X]$ with a high probability.}
\label{fig:mitigation_only}
\end{figure}

Since larger $|W_{avg}|$ can control the difference of estimated perturbations between consecutive steps (i.e.  $\Delta$ $\hat{\epsilon}_{j, j-1}$ = $\hat{\epsilon}_{j-1}$ - $\hat{\epsilon}_{j}$) in a finer-granule manner, Algorithm \ref{alg:mitigating_perturbation} with a large $|W_{avg}|$ can get $X_{adv}$ closer to $X$ so that it achieves a highier prediction accuracy than Algorithm \ref{alg:mitigating_perturbation} with the smaller $|W_{avg}|$.
Figure \ref{fig:mitigation_only} shows that Algorithm \ref{alg:mitigating_perturbation} with a large $|W_{avg}|$ gets the high prediction accuracy for the adversarial example having a large perturbation (i.e. $\epsilon$ $=$ 64) when $Pr[Y_{true}|X]$ $\approx$ $Pr[Y_{true}|W_{avg}*X]$ with a high probability.

\begin{figure}[htb]
\begin{subfigure}[htb]{0.5\textwidth}
  \includegraphics[width=\textwidth]{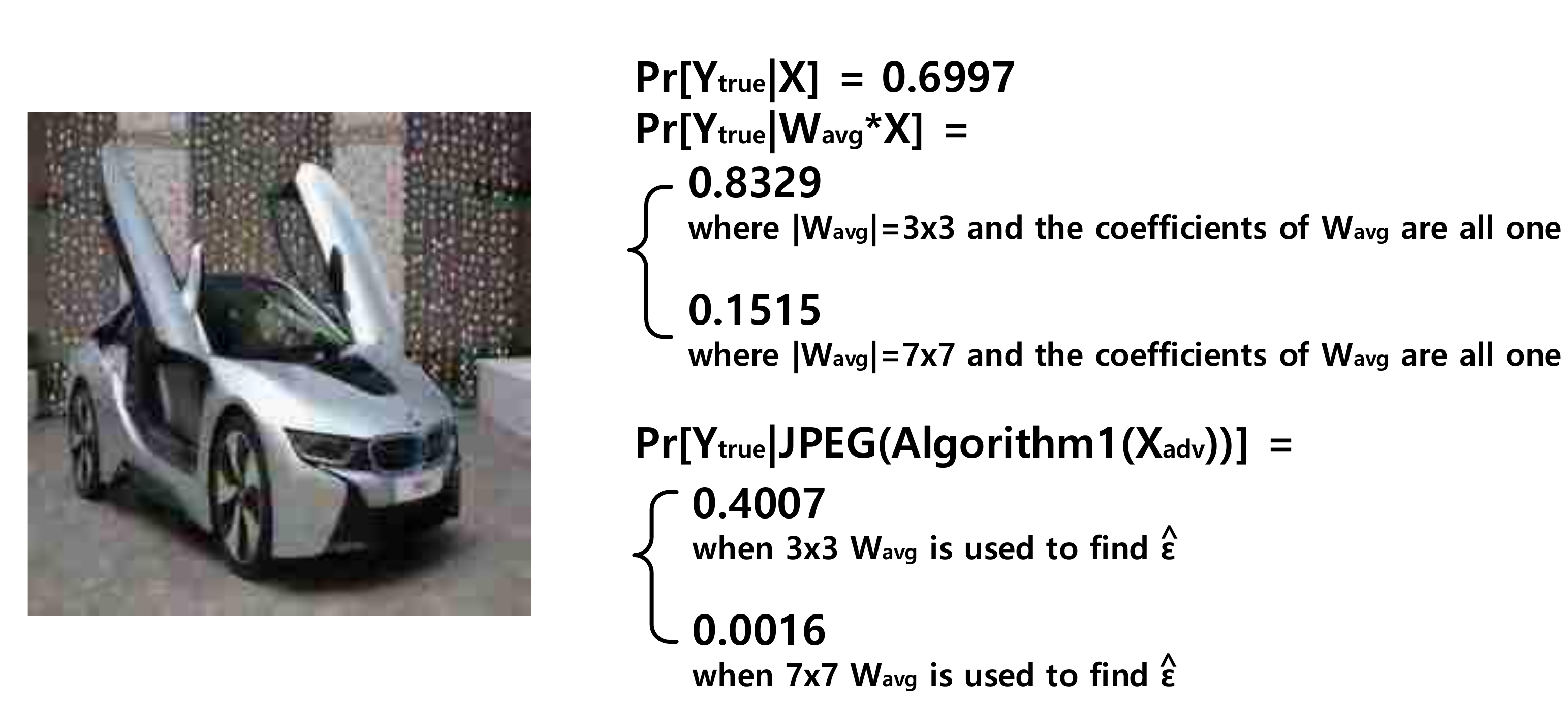}
	\caption{Algorithm \ref{alg:mitigating_perturbation} cannot improve the prediction accuracy of the adversarial example $X_{adv}$ (generated by the basic iterative FGSM attack with $\epsilon$ $=$ 64) very much when $Pr[Y_{true}|X]$ $\not\approx$ $Pr[Y_{true}|W_{avg}*X]$.}
  \label{fig:sports_car_low_prob}
\end{subfigure}
\begin{subfigure}[htb]{0.5\textwidth}
  \includegraphics[width=\textwidth]{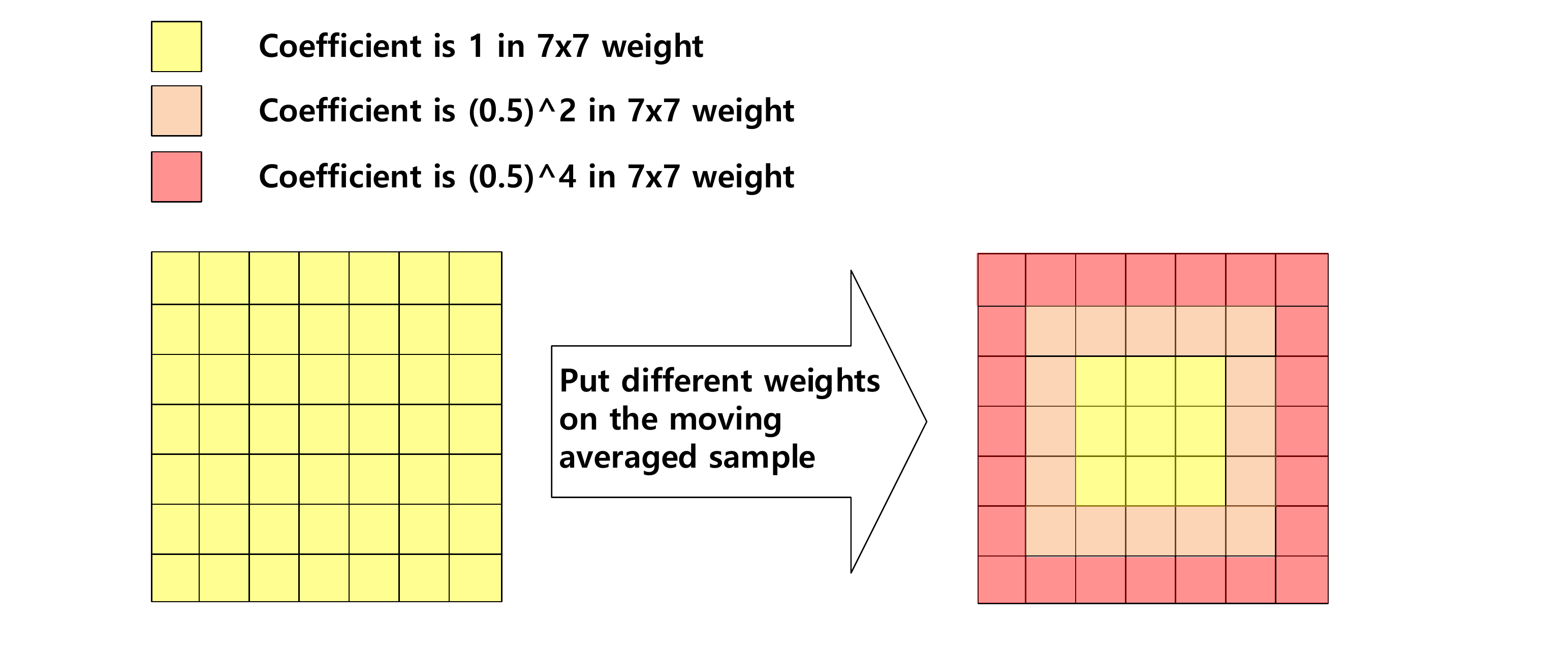}
  \caption{Changing the coefficients of $W_{avg}$ in Figure \ref{fig:sports_car_low_prob}  
    by putting more weights on the coefficients closer to the centroid of $W_{avg}$}
  \label{fig:change_coefficients}
\end{subfigure}
\begin{subfigure}[htb]{0.5\textwidth}
  \includegraphics[width=\textwidth]{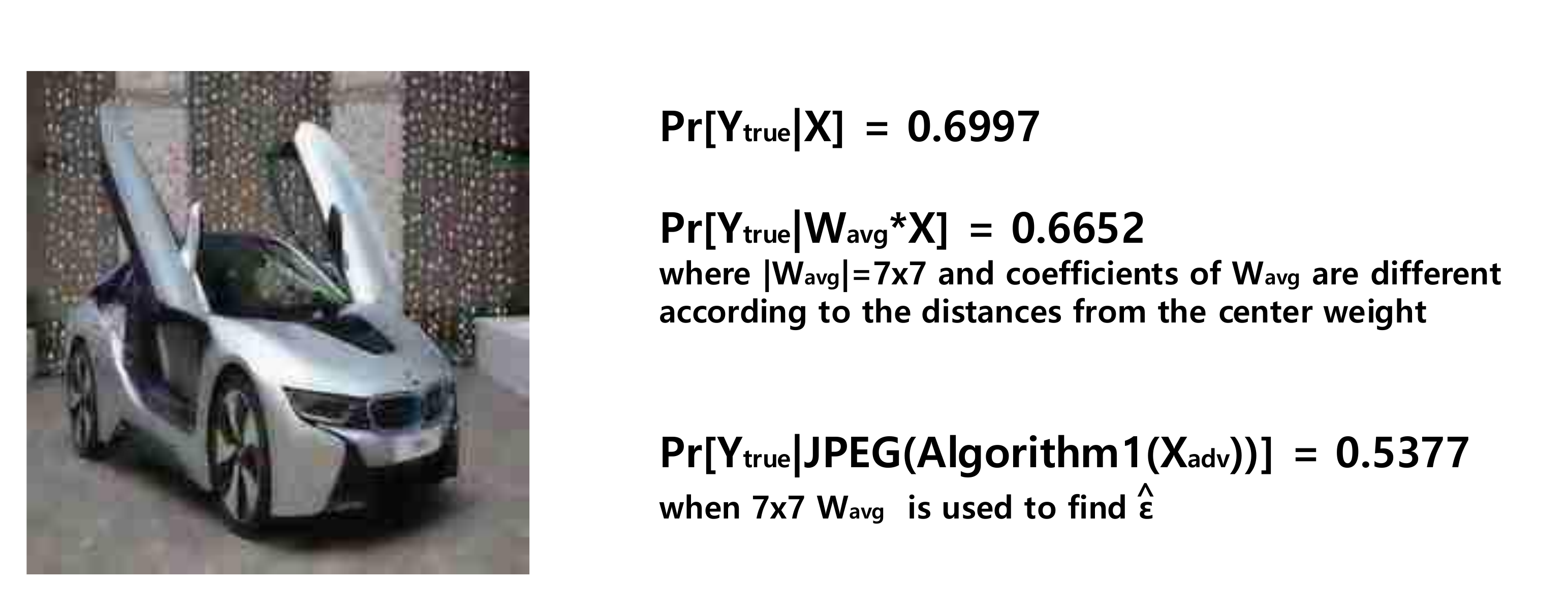}
	\caption{Algorithm \ref{alg:mitigating_perturbation} works well on the adversarial example $X_{adv}$ (generated by the basic iterative FGSM attack with $\epsilon$ $=$ 64) after the coefficients of $W_{avg}$ is changed to satisfy $Pr[Y_{true}|X]$ $\approx$ $Pr[Y_{true}|W_{avg}*X]$.}
  \label{fig:sports_car_high_prob}
\end{subfigure}

\caption{Changing the coefficients of $W_{avg}$ can increase the prediction accuracy of the adversarial example mitigated by  Algorithm \ref{alg:mitigating_perturbation} when $Pr[Y_{true}|X]$ $\not\approx$ $Pr[Y_{true}|W_{avg}*X]$.}
\label{fig:weight_changes}
\end{figure}

In Figure \ref{fig:mitigation_only}, both Figure \ref{fig:miti_ambulance_with_large_W} and \ref{fig:miti_ambulance_with_small_W} satisfy $Pr[Y_{true}|X]$ $\approx$ $Pr[Y_{true}|W_{avg}*X]$ with a high probability.
However, Algorithm \ref{alg:mitigating_perturbation} mitigates Figure \ref{fig:miti_ambulance_with_large_W} better than Figure \ref{fig:miti_ambulance_with_small_W}.
In Figure \ref{fig:miti_ambulance_with_large_W}, Algorithm \ref{alg:mitigating_perturbation} can get some high prediction accuracy even without JPEG encoding ($Pr[Y_{true}|Algorithm1(X_{adv})]$ $=$ 0.6118) but it cannot get any prediction accuracy ($Pr[Y_{true}|Algorithm1(X_{adv})]$ $=$ 0.0000) in Figure \ref{fig:miti_ambulance_with_small_W}.
In case that $Pr[Y_{true}|X]$ $\not\approx$ $Pr[Y_{true}|W_{avg}*X]$ and both prediction accuracies are not so high,
we can improve the prediction accuracy of the adversarial example mitigated by Algorithm \ref{alg:mitigating_perturbation} by changing the coefficients of $W_{avg}$. 
Figure \ref{fig:weight_changes} shows that 
changing the coefficients of $W_{avg}$ can improve the prediction accuracy of the adversarial example mitigated by Algorithm \ref{alg:mitigating_perturbation} when $Pr[Y_{true}|X]$ $\not\approx$ $Pr[Y_{true}|W_{avg}*X]$.

In Figure \ref{fig:sports_car_low_prob}, $Pr[Y_{true}|X]$ $\not\approx$ $Pr[Y_{true}|W_{avg}*X]$ when the larger (i.e. $7\times7$) $W_{avg}$ is used to find $\hat{\epsilon}$. In order to make $Pr[Y_{true}|X]$ $\approx$ $Pr[Y_{true}|W_{avg}*X]$ for $7\times7$ $W_{avg}$, $7\times7$ $W_{avg}$ needs to be changed to have the coefficients which are similar to $3\times3$ $W_{avg}$.
To get the same result with $3\times3$ $W_{avg}$ by using $7\times7$ $W_{avg}$, $3\times3$ coefficients of the centroid in $7\times7$ $W_{avg}$ are filled with ones and other coefficents are set as zeros.
Thus, in order to make the prediction accuracy of $W_{avg}*X$ with $|W_{avg}|$ $=$ $7\times7$ ($=$0.1515) grow toward that of $W_{avg}*X$ with $|W_{avg}|$ $=$ $3\times3$ ($=$0.8329). in Figure \ref{fig:change_coefficients}, $3\times3$ coefficients of the centroid in $7\times7$ $W_{avg}$ are filled with ones and other coefficients except the $3\times3$ coefficients are set as small values (i.e. $<$ 1) as the distance from the centroid increases.
As the result, in Figure \ref{fig:sports_car_high_prob}, $Pr[Y_{true}|X]$ $\approx$ $Pr[Y_{true}|W_{avg}*X]$ and Algorithm \ref{alg:mitigating_perturbation} mitigates $X_{adv}$ very well to have the good prediction accuracy (i.e. $Pr[Y_{true}|JPEG(Algorithm1(X_{adv}))]$ $=$ 0.5377).
However, we do not know the size of the $W_{avg}$ (where all coefficients are ones) to make $Pr[Y_{true}|X]$ $\approx$ $Pr[Y_{true}|W_{avg}*X]$ as well as which coefficients are the best suited for the mitigation of $X_{adv}$ when $Pr[Y_{true}|X]$ $\not\approx$ $Pr[Y_{true}|W_{avg}*X]$ because $X_{adv}$ is just given for the mitigation process (corresponding to Algorithm \ref{alg:mitigating_perturbation} in this paper). Therefore, the ways to make $Pr[Y_{true}|X]$ $\approx$ $Pr[Y_{true}|W_{avg}*X]$ for any given $X_{adv}$ need to be further studied.


%
%
%
%

\end{document}